\documentclass[final,5p]{elsarticle}
\usepackage{array}
\usepackage{times}
\usepackage{graphicx}
\usepackage{psfrag}
\usepackage{epsfig}
\usepackage{algorithm}
\usepackage{algorithmic}
\usepackage{multicol}
\usepackage{multirow}
\usepackage{setspace}
\usepackage{array}
\usepackage{amssymb}
\usepackage[cmex10]{amsmath}
\usepackage{mdwmath}
\usepackage{mdwtab}
\usepackage[tight,footnotesize]{subfigure}
\usepackage{mdwlist}
\usepackage{caption}
\usepackage{float}
\usepackage{slashbox}
\usepackage[amssymb]{SIunits}
\usepackage[usenames]{color}
\usepackage{bm}
\usepackage{cases}
\usepackage{tabularx}
\usepackage[colorlinks,linkcolor=black,anchorcolor=black,citecolor=black,urlcolor=black]{hyperref}

\usepackage{amssymb}

\journal{Pattern Recognition}

\begin{document}

\begin{frontmatter}



\title{KCRC-LCD: Discriminative Kernel Collaborative Representation with Locality Constrained Dictionary for Visual Categorization}


\author[label1]{Weiyang Liu}
\author[label2]{Zhiding Yu\corref{mycorrespondingauthor}}
\cortext[mycorrespondingauthor]{Corresponding author}
\ead{yzhiding@andrew.cmu.edu}

\author[label1]{Lijia Lu}
\author[label3]{Yandong Wen}
\author[label1]{Hui Li}
\author[label1]{Yuexian Zou}

\address[label1]{School of Electronic and Computer Engineering, Peking University, China}
\address[label2]{Department of Electrical and Computer Engineering, Carnegie Mellon University, U.S.}
\address[label3]{School of Electronic and Information Engineering, South China University of Technology, China}
\begin{abstract}
We consider the image classification problem via kernel collaborative representation classification with locality constrained dictionary (KCRC-LCD). Specifically, we propose a kernel collaborative representation classification (KCRC) approach in which kernel method is used to improve the discrimination ability of collaborative representation classification (CRC). We then measure the similarities between the query and atoms in the global dictionary in order to construct a locality constrained dictionary (LCD) for KCRC. In addition, we discuss several similarity measure approaches in LCD and further present a simple yet effective unified similarity measure whose superiority is validated in experiments. There are several appealing aspects associated with LCD. First, LCD can be nicely incorporated under the framework of KCRC. The LCD similarity measure can be kernelized under KCRC, which theoretically links CRC and LCD under the kernel method. Second, KCRC-LCD becomes more scalable to both the training set size and the feature dimension. Example shows that KCRC is able to perfectly classify data with certain distribution, while conventional CRC fails completely. Comprehensive experiments on many public datasets also show that KCRC-LCD is a robust discriminative classifier with both excellent performance and good scalability, being comparable or outperforming many other state-of-the-art approaches.
\end{abstract}


\end{frontmatter}


\section{Introduction}
Recent years have witnessed the great success of the sparse representation techniques in a variety of problems in computer vision, including image restoration \cite{mairal2010online}, image denoising \cite{elad2006image} as well as image classification \cite{wright2009robust,yang2009linear,yuan2012visual}. Sparse representation is widely believed to bring many benefits to classification problems in terms of robustness and discriminativeness.
Specifically, sparsity is a regularizer that can reduce the solution space under ill-conditioned problems, by seeking to represent a signal as a linear combination of only a few bases. These bases are called the ``atoms'' and the whole overcomplete collection of atoms together form what one call a ``dictionary''. Many natural signals such as image and audio indeed have sparse priors. Imposing sparsity not only returns a unique solution, but also helps to recover the true signal structure, giving more robust estimation against noise. In addition, the sparse representation of a signal often leads to better separation and decorrelation which benefits subsequent classification problems. Despite the fact that sparse optimization is a nonconvex problem, the $l_{1}$-norm convex relaxation and its optimization techniques have been thoroughly studied.
\par
Wright et al. \cite{wright2009robust} employed the entire set of the training samples as the dictionary and reported a discriminative sparse representation-based classification (SRC) with promising performance on face recognition. SRC approximates an input signal $\bm{y}$ with a linear combination of the atoms from an overcomplete dictionary $\bm{D}$ under the sparsity constraint and gives the predicted label by selecting the minimum reconstruction residuals. Despite the fact that SRC was widely used in various applications \cite{lang2012saliency,guha2012learning,mei2011robust}, \cite{rigamonti2011sparse,shi2011face,zhang2011sparse} still questioned the role that sparse representation plays in the image classification tasks.
\par
Zhang et al. \cite{zhang2011sparse} further commented that it is unnecessary to enforce the sparse constraint with computationally expensive $l_1$-norm if the feature dimension is high enough. Their work emphasized the importance of collaborative representation (CR) rather than the sparse representation, arguing that CR is the key to the improvement of classification accuracy, which was validated by their comparison experiments. They used the $l_2$-norm regularization instead of $l_1$-norm, further improving the classification accuracy while significantly reducing much computational cost. The corresponding proposed method is called collaborative representation classification (CRC).
\par
Despite their robust performance, the linear nature of both SRC and CRC makes them perform poorly when the training data are distributed vector-like in one direction. Kernel function, proven useful in kernel principle component analysis (KPCA) \cite{scholkopf1997kernel} and support vector machine (SVM) \cite{burges1998tutorial}, was introduced to overcome such shortcoming for both SRC and CRC, leading to the kernel sparse representation-based classification (KSRC) \cite{zhang2012kernel} and the kernel collaborative representation classification (KCRC) \cite{liu2014novel}. In particular, a Mercer kernel implicitly defines a nonlinear mapping to map the data from the input space into a high or even infinite dimensional kernel space where features of the same class can be grouped together more easily and different classes become linear separable. 
\par
Besides summarizing the KCRC \cite{liu2014novel}, our major contribution in this paper lies in proposing a generalized framework for KCRC with locality constrained dictionary and unified similarity measure, giving both performance gain and significant reduction of computational cost. Due to the poor scalability of the global dictionary (GD) used in CR-based methods, classification becomes intractable in large database for KCRC with GD (KCRC-GD). To enable the scalability to large databases, we prune the dictionary via $k$-nearest neighbor (K-NN) classifier to enforce locality. Specifically, the nearest neighbors of a query serve to construct a locality constrained dictionary (LCD) for KCRC. Such strategy is both intuitively reasonable and mathematically appealing. Intuitively, LCD is well motivated by the psychological findings about human perception that visual categories are not defined by lists of features, but rather by similarity to prototypes \cite{Rosch1976basic}. In other word, coarse level matching, for which K-NN is used, plays an important role in human perception. Mathematically, LCD can be nicely incorporated under the framework of KCRC. First, the LCD similarity measure can also be kernelized under KCRC, which theoretically links CRC and LCD under the kernel method. Second, KCRC-LCD becomes more scalable to both the training set size and the feature dimension. The kernel gram matrix is now obtained from a subset of the global dictionary, while KCRC operates on the reduced kernel matrix without referring to original features.
\par
The high level intuition of LCD is that local dictionary atoms are typically the most important and informative samples. Looking into these representative exemplars often brings even more gains than globally considering all samples together. It is not hard to see similar concepts and link the connections. For example, if the query is located near decision boundary, then these local atoms play the role similar to support vectors in an SVM, or in an extreme case, exemplars in an exemplar SVM \cite{malisiewicz2011ensemble}. In a model recommendation system, selecting the most responsive models instead of all models has been reported to give gains [15]. In fact, SRC also seeks to use only few exemplar atoms in the dictionary. Yet the proposed method is able to run much faster with even better performance. In the extreme scenario, if K equals the number of atoms, then the proposed KCRC-LCD degenerates to regular KCRC with global dictionary. If $K$ equals 1, KCRC-LCD degenerates to the simplest nearest neighbor classifier.

In this paper, we specifically focus on the application of our proposed framework to the image classification/visual categorization problem, demonstrating its robust performance with a comprehensive series of image classification tasks. Image classification is among the most fundamental computer vision problems where each image is labeled with a certain or multiple categories/tags. Though great advance has been achieved, much is pending to be done since the current state-of-the-art approaches are far from being able to achieve human-level performance, particularly in handling cluttered, complicated scenarios and inferring abstract concepts. Such gap between the machine and human remains an open challenge, motivating us to exploit more discriminative and efficient image classifiers. 

\par
The outline of the paper is as follows. Section \uppercase\expandafter{\romannumeral2} discusses related work of KCRC-LCD and presents our main contributions. In Section \uppercase\expandafter{\romannumeral3}, necessary preliminaries are briefly introduced. Section \uppercase\expandafter{\romannumeral4} elaborates the formulation of KCRC-LCD and discusses some important details. The locality constrained dictionary is proposed and discussed in Section \uppercase\expandafter{\romannumeral5}. Experimental results are provided and discussed in Section \uppercase\expandafter{\romannumeral6}, followed by concluding remarks in Section \uppercase\expandafter{\romannumeral7}.
\section{Related Work}
Pioneering work for combining kernel technique to SRC is proposed in \cite{gao2010kernel,zhang2012kernel,yin2012kernel}. Gao et al. first proposed the idea of kernel-based SRC with a promising experimental results. Zhang et al. \cite{zhang2012kernel} further unified the mathematical model \cite{gao2010kernel} to a generic kernel framework and conducted more comprehensive experiments to evaluate the performance. To overcome the shortcoming of handling data with the same direction distribution, Liu et al. \cite{liu2014novel} addressed the problem of kernel collaborative representation. The authors presented a smooth formulation to incorporate kernel function into the CRC model. A practical application of kernel CRC in vehicle logo recognition was further discussed in \cite{liu2014kernel}. We happened to notice that a very recent work \cite{li2014column} proposed a similar idea by combining the column-generation kernel to CRC for hyperspectral image classification. It should be pointed out, however, that both the formulations as well as the applications are significantly different when dealing with the high dimension in kernel space. In terms of application, \cite{li2014column} formulated the kernel collaborative representation on pixel-level tasks for hyperspectral images while our method focus on image-level classification. Significant differences also exist in the formulation: \cite{li2014column} incorporated the kernel function with column generation without considering dimensionality reduction in kernel space (possibly due to the characteristics of the hyperspectral classification task). On the contrary, our method combines the CRC with kernel function in a strategy similar to KPCA and kernel Fisher discriminant analysis (KFDA). Moreover, a series of dimensionality reduction approaches have been taken into account in our generalized formulation. In general, we aim at extending the idea of KCRC by further improving formulation details of KCRC and presenting specific methods to perform dimensionality reduction in kernel space.
\par
We also noticed that \cite{li2014hyperspectral} presented a similar idea of constructing locally adaptive dictionary, but such dictionary pruning strategy was only applied in the standard CRC framework instead of the kernel CRC framework. As one shall see, the proposed KCRC-LCD is not a trivial extension of CRC-LCD by combining KCRC with locality constrained dictionary, but a well-motivated and appealing framework in which LCD and KCRC are theoretically linked by kernelizing the distance used in LCD. In addition, the kernelization in conjunction with LCD not only brings scalability in terms of data set size, but also further extends its scalability to feature dimensionality. We will show that the locally adaptive dictionary in \cite{li2014hyperspectral} is a special case of our proposed LCD. 
\section{Preliminaries}
\subsection{Collaborative Representation Classification}
The principle of CRC \cite{zhang2011sparse} is briefly presented in this section. In CRC, the dictionary $\bm{D}$ is constructed by all training samples and a test sample $\bm{y}$ is coded collaboratively over the whole dictionary, which is the essence of CRC.
\par
Let $\bm{D}$ be the dictionary that is a set of $k$-class training samples (the $i$th class with $n_i$ samples), i.e., $\thickmuskip=0mu \bm{D}=\{  \bm{D}_1,\bm{D}_2,\cdots,\bm{D}_k \}\in \mathbb{R}^{m \times n}$ where $\thickmuskip=0mu n=\sum_{j=1}^k n_j$ and $m$ is the feature dimension. Here, the dictionary associated with the $i$th class is denoted by $\thickmuskip=0mu \bm{D}_i=\{\bm{d}_{1}^{[i]},\bm{d}_2^{[i]},\cdots,\bm{d}_{n_i}^{[i]}\}\in \mathbb{R}^{m \times n_i}$ in which $\bm{d}_{j}^{[i]}$ - also called atom - stands for the $j$th training image in the $i$th class. Representing the query sample $\bm{y}$ can be accomplished by solving $\bm{x}$ in the following optimization model:
\begin{equation}\label{egn:CRC}
\begin{aligned}
&\ \ \ \ \hat{\bm{x}}=\arg \min_{\bm{x}}\| \bm{x} \|_{l_p} \\
&\text{\textnormal{subj. to}} \ \| \bm{y}- \bm{D}\bm{x} \|_{l_q} \leq \varepsilon
\end{aligned}
\end{equation}
where $\varepsilon$ is a small error constant. After Lagrangian formulation, the model of CRC can be formulated as
\begin{equation}\label{egn:CRC2}
\hat{\bm{x}}=\arg \min_{\bm{x}}\big( \| \bm{y}- \bm{D}\bm{x} \|_{l_q} +\mu\| \bm{x} \|_{l_p} \big)
\end{equation}
where $\mu$ is the regularization parameter and $p,q\in\{1,2\}$. The combinations of $p,q$ lead to different instantiations of CRC model. For instance, SRC method is under the condition of $\thickmuskip=0mu p=1,q\in\{1,2\}$ and different settings of $q$ are used to handle classification with or without occlusion.
Similar to SRC, CRC determines the class label via reconstruction residuals:
\begin{equation}\label{egn:res}
identity(\bm{y})=\arg \min_{i}\big( \| \bm{y}- \bm{D}_i\hat{\bm{x}}_i \|_{2} / \| \hat{\bm{x}}_i \|_{2} \big).
\end{equation}
\par
In fact, the key to reduce the computational complexity is to reasonably set the value of $p,q$. Based on different combinations of $p,q$, two CRC algorithms were proposed in \cite{zhang2011sparse,zhang2012CRC}. One is the CRC regularized least square (CRC-RLS) algorithm with $\thickmuskip=0mu p=2,q=2$. The other is the robust CRC (RCRC) algorithm with $\thickmuskip=0mu p=2,q=1$. \cite{zhang2011sparse} concluded that sparsity of signals is useful but not crucial for face recognition and proved that the collaborative representation mechanism does play an important role.
\subsection{Kernel Technique}
In machine learning, kernel methods refer to a class of algorithms for pattern analysis, whose best known members are the SVM \cite{vapnik1998statistical,burges1998tutorial}, KPCA \cite{scholkopf1997kernel} and KFDA \cite{mika1999fisher}. The general task of pattern analysis is to find and study general types of relations (for example clusters, rankings, principal components, correlations, classifications) in datasets. For many algorithms that solve these tasks, the data in raw representation have to be explicitly transformed into feature vector representations via a user-specified feature map: in contrast, kernel methods require only a user-specified kernel, i.e., a similarity function over pairs of data points in raw representation.  Via kernels, we can easily generalize a linear classifier to a nonlinear one, generating a reasonable decision boundary and consequently enhancing the discrimination power.
\par
Kernel methods owe their name to the use of kernel functions, which enable them to operate in a high-dimensional, implicit feature space without computing the coordinates of the data in that space, but rather by simply computing the inner products between the images of all pairs of data in the feature space. This operation is often computationally cheaper than the explicit computation of the coordinates. Kernel functions have been introduced for sequence data, graphs, text, images, as well as vectors. The amazing part of kernel function is that it surpasses the direct calculation in the feature space and performs the classification in the reproducing kernel Hilbert space (RKHS), boosting the classification performance.
\par
Algorithms that are capable of operating with kernels include the kernel perceptron \cite{orabona2008projectron,he2012kernel,dekel2008forgetron}, SVM \cite{vapnik1998statistical,burges1998tutorial}, Gaussian processes \cite{rasmussen2006gaussian}, principal components analysis (PCA) \cite{scholkopf1997kernel}, canonical correlation analysis \cite{hardoon2004canonical}, spectral clustering \cite{ng2002spectral} and many others. In general, any linear model can be transformed into a non-linear model by applying the kernel trick: replacing its features by a kernel function.
\section{Proposed KCRC Approach}
\subsection{Formulation of KCRC}
To overcome the shortcoming of CRC in handling data with the same direction distribution, kernel technique is smoothly combined with CRC. Kernel function is used to create a nonlinear mapping mechanism $\thickmuskip=0.5mu \bm{v}\in\mathbb{R}\mapsto\bm{\phi}(\bm{v})\in\mathbb{H}$ in which $\mathbb{H}$ is a unique associated RKHS. If every sample is mapped into higher dimensional space via transformation $\bm{\phi}$, the kernel function is written as
\begin{equation}\label{egn:Kproduct}
K(\bm{v}',\bm{v}'')=\langle\bm{\phi}(\bm{v}'),\bm{\phi}(\bm{v}'')\rangle=\bm{\phi} (\bm{v}')^T\bm{\phi}(\bm{v}'')
\end{equation}
where $\bm{v}',\bm{v}''$ are different samples and $\bm{\phi}$ denotes the implicit nonlinear mapping associated with the kernel function $K(\bm{v}',\bm{v}'')$. There are some empirical kernel functions satisfying the Mercer condition such as the linear kernel $\thickmuskip=0mu K(\bm{v}',\bm{v}'')=\bm{v}'^T\bm{v}''$ and radial basis function (RBF) kernel $\thickmuskip=0mu K(\bm{v}',\bm{v}'')=\text{\textnormal{exp}}(-\beta\|\bm{v}'-\bm{v}''\|_2^2)$. According to \cite{scholkopf2001kernel}, the distance function for similarity measurement, designed to construct the LCD, can be transformed in a straightforward way to the kernel for KCRC, via the linear kernel function:
\begin{equation}\label{egn:Knew}
\begin{aligned}
 K&(\bm{v}',\bm{v}'')  \medmuskip=1mu =\langle \bm{\phi}(\bm{v}),\bm{\phi}(\bm{v}')\rangle=\langle \bm{v},\bm{v}'\rangle\\
  &\medmuskip=1mu =\frac{1}{2}\big(\langle \bm{v}',\bm{v}'\rangle+\langle \bm{v}'',\bm{v}''\rangle-\langle \bm{v}'-\bm{v}'',\bm{v}'-\bm{v}''\rangle\big)\\
  &\medmuskip=1mu =\frac{1}{2}\big(Dist(\bm{v}',0)+Dist(\bm{v}'',0)-Dist(\bm{v}',\bm{v}'')\big)
\end{aligned}
\end{equation}
where $Dist$ is the carefully designed distance function, and the location of the origin(0) does not affect the result \cite{scholkopf2001kernel}. Various ways of transforming a distance function into a kernel are possible \cite{zhang2006svm}, i.e., $K(\bm{v}',\bm{v}'')$ can be $\text{\textnormal{exp}}(-Dist(\bm{v}',\bm{v}'')/\beta^2)$.
\par
It is learned in \cite{burges1998tutorial} that the sample feature nonlinearly transformed to high dimensional space becomes more separable. Most importantly, the same direction distribution of data can be avoided in kernel space. However, mapping to high dimensional space makes CRC model harder to solve, so we need to perform dimensionality reduction in the kernel feature space.
The nonlinear mapping mechanism is
\begin{equation}\label{egn:KMap}
\bm{y}\in\mathbb{R}^m\mapsto\bm{\phi}(\bm{y})=[\phi_1(\bm{y}),\phi_2(\bm{y}),\cdots,\phi_s(\bm{y})]\in\mathbb{R}^s
\end{equation}
where $\bm{\phi}(\bm{y})$ is the high dimensional feature corresponding to the sample $\bm{y}$ and $\thickmuskip=0.5mu s\gg m$. Then we define a universal label $[k]$ for $\bm{d}_j^{[i]}$ that denotes its position in the global dictionary, satisfying $\thickmuskip=1mu \medmuskip=1mu k=j+\sum_{l=1}^{i-1}n_l$. For conciseness, we only preserve the universal label, representing atom as $\bm{d}_{[k]}$. According to the nonlinear mapping mechanism, the original dictionary $\bm{D}$ has a much higher dimension: $\thickmuskip=0.5mu \bm{\Phi}=\{ \bm{\phi}(\bm{d}_{[1]}) , \bm{\phi}(\bm{d}_{[2]}) , \cdots , \bm{\phi}(\bm{d}_{[n]}) \}\in\mathbb{R}^{s\times n}$, and the test sample becomes $\thickmuskip=0mu \bm{\phi}(\bm{y})=\bm{\Phi}\bm{x}$. The KCRC model is formulated as
\begin{equation}\label{egn:KL1}
\begin{aligned}
&\ \ \ \ \ \ \ \hat{\bm{x}}=\arg\min_{\bm{x}} \| \bm{x} \|_{l_p} \\
&\text{\textnormal{subj.\ to}} \ \|\bm{\phi}(\bm{y})-\bm{\Phi}\bm{x}\|_{l_q}\leq \varepsilon \ldotp
\end{aligned}
\end{equation}
However, Eq. \eqref{egn:KL1} is harder to solve than Eq. \eqref{egn:CRC}
because the high dimensionality results in high complexity. A dimensionality reduction matrix $\bm{R}$, namely a projection matrix, can be constructed by utilizing the methodology in KPCA \cite{scholkopf1997kernel} and KFDA \cite{mika1999fisher}. With the matrix $\bm{R}\in\mathbb{R}^{s\times c}$, we derive
\begin{equation}\label{egn:MatR}
\bm{R}^T\bm{\phi}(\bm{y})=\bm{R}^T\bm{\Phi}\bm{x}
\end{equation}
where $\bm{R}$ is related to kernelized samples. In both KPCA and KFDA, each column vector in $\bm{R}$ is a linear combination of kernelized samples, which is also adopted in KCRC. Namely
\begin{equation}\label{egn:MatRDef}
\bm{R}=\bm{\Phi}\bm{\Psi}=\{\bm{\phi}(\bm{d}_{[1]}),\cdots,\bm{\phi}(\bm{d}_{[n]})\}\cdot\{\bm{\psi}_1,\cdots,\bm{\psi}_c\}
\end{equation}
where $\thickmuskip=0.5mu \bm{R}=\{\bm{R}_1,\cdots,\bm{R}_s\}$ and $\bm{\psi}_i$ is the $n$-dimensional linear projection coefficients vector: $\thickmuskip=0.5mu \bm{R}_i=\sum_{j=1}^n\psi_{i,j}\bm{\phi}(\bm{d}_{[j]})=\bm{\Phi}\bm{\psi}_i$. Moreover, $\thickmuskip=0.5mu \bm{\Psi}\in\mathbb{R}^{n\times c}$ is also called pseudo-transformation matrix \cite{zhang2012kernel}. Then we put Eq. \eqref{egn:MatRDef} into Eq. \eqref{egn:MatR} and obtain
\begin{equation}\label{egn:KRep}
(\bm{\Phi}\bm{\Psi})^T\bm{\phi}(\bm{y})=(\bm{\Phi}\bm{\Psi})^T\bm{\Phi}\bm{x}
\end{equation}
from which we get $\thickmuskip=0mu \bm{\Psi}^T \bm{K}(\bm{D},\bm{y})=\bm{\Psi}^T\bm{G}\bm{x}$, where $\thickmuskip=0mu \bm{K}(\bm{D},\bm{y})=[K(\bm{d}_{[1]},\bm{y}),\cdots,K(\bm{d}_{[n]},\bm{y})]^T$. $\bm{G}$ ($\thickmuskip=0mu G_{ij}=K(\bm{d}_{[i]},\bm{d}_{[j]})$), also equal to $\bm{\Phi}^T\bm{\Phi}$, is defined as the kernel Gram matrix that is symmetric and positive semi-definite according to Mercer's theorem. Since $\bm{G}$ and $\bm{K}(\bm{D},\bm{y})$ are given a prior, dimensionality reduction requires to find $\bm{\Psi}$ instead of $\bm{R}$. Several methods were introduced in \cite{zhang2012kernel,mika1999fisher,scholkopf1997kernel} to determine the pseudo transformation matrix $\bm{\Psi}$. We will also further introduce the selection of matrix $\bm{\Psi}$ in the next subsection. Note that, if $\bm{\Psi}$ is an identity matrix, no dimensionality reduction is applied. Particularly, $\bm{\Psi}$ can also be a random projection matrix to achieve dimensionality reduction. After substituting the equivalent kernel function constraint, we can derive
\begin{equation}\label{egn:KOpt}
\begin{aligned}
&\ \ \ \ \ \ \ \ \ \ \ \ \  \hat{\bm{x}}=\arg\min_{\bm{x}} \| \bm{x} \|_{l_p} \\
&\text{\textnormal{subj. to}} \ \| \bm{\Psi}^T \bm{K}(\bm{D},\bm{y})-\bm{\Psi}^T\bm{G}\bm{x} \|_{l_q}<\varepsilon
\end{aligned}
\end{equation}
which is the model of KCRC approach. Additionally, a small perturbation would be added to $\bm{\Psi}^T\bm{G}$ if the norm of a column is close to 0. Another form of KCRC model is expressed as
\begin{equation}\label{egn:KOpt2}
\hat{\bm{x}}=\arg\min_{\bm{x}} \big(\| \bm{\Psi}^T \bm{K}(\bm{D},\bm{y})-\bm{\Psi}^T\bm{G}\bm{x} \|_{l_q}+ \mu\| \bm{x} \|_{l_p} \big)
\end{equation}
from which we could derive two specific algorithms. With $\thickmuskip=0mu p=2,q=2$, $\bm{x}$ can be solved at the cost of low computational complexity. The regularized least square algorithm is used to solve the optimization problem (Algorithm 1).
Handling images with occlusion and corruption, we can set $\thickmuskip=0mu p=2,q=1$ for robustness, making the first term a $l_1$ regularized one. Let $\thickmuskip=0mu \bm{e}=\bm{\Psi}^T \bm{K}(\bm{D},\bm{y})-\bm{\Psi}^T\bm{G}\bm{x}$ and $\thickmuskip=0mu p=2,q=1$. Eq. \eqref{egn:KOpt} is rewritten as
\begin{equation}\label{egn:KRCRC}
\begin{aligned}
&\ \ \ \ \hat{\bm{x}}=\arg\min_{\bm{x}} \big(\| \bm{e} \|_{1}+ \mu\| \bm{x} \|_{2} \big)\\
&\text{\textnormal{subj.\ to}} \ \ \bm{\Psi}^T \bm{K}(\bm{D},\bm{y})=\bm{\Psi}^T\bm{G}\bm{x}+\bm{e}
\end{aligned}
\end{equation}
which is a constrained convex optimization problem that can be solved by the augmented Lagrange multiplier (ALM) method \cite{bertsekas1999nonlinear,lin2010augmented} as shown in Algorithm 2.
\subsection{Determining the Pseudo-transformation Matrix for Dimensionality Reduction}
This subsection reviews several typical methods that are proposed in \cite{zhang2012kernel} to determine the pseudo-transformation matrix $\bm{\Psi}$ for dimensionality reduction. Moreover, we also present a graph preserving method that has not been utilized to construct pseudo-transformation matrix in previous work.
\subsubsection{KPCA}
Following the methodology in KPCA, the pseudo-transformation vectors $\bm{\psi_i}\in\mathbb{R}^n$ refer to normalized eigenvectors corresponding to nonzero eigenvalues (or greater than a threshold) which can be obtained from the following eigenvalue problem \cite{scholkopf1997kernel}:
\begin{equation}\label{egn:EigenVal}
n\lambda\bm{\psi}_i=\bm{G}\bm{\psi}_i
\end{equation}
where $\bm{\psi}_i$ is normalized to satisfy $\lambda_i\bm{\psi}_i^T\bm{\psi}_i=1$. Eq. \ref{egn:EigenVal} can be easily solved by singular value decomposition (SVD) method. $\bm{\Psi}$ is equal to $\{\bm{\psi}_1,\bm{\psi}_2,\cdots,\bm{\psi}_c\}$.
\subsubsection{KFDA}
For KFDA \cite{mika1999fisher}, $\bm{\Psi}\in\mathbb{R}^{n\times c}$ is the solution of the optimization problem shown as follows:
\begin{equation}\label{egn:KFDAopt}
\arg\max_{\bm{\Psi}}\ \dfrac{tr\big(\bm{\Psi}\bm{S}_b^{\bm{G}}\bm{\Psi}\big)}{tr\big(\bm{\Psi}\bm{S}_\omega^{\bm{G}}\bm{\Psi}\big)}
\end{equation}
where $tr(\cdot)$ denotes the trace of a matrix, and $\bm{S}_b^{\bm{G}},\bm{S}_\omega^{\bm{G}}$ stand for quasi within-class and between-class scatter matrices respectively.
\subsubsection{Random Projection}
\cite{zhang2012kernel} also proposed a simple and practical random dimensionality reduction method. Since random projection can not be performed in the RHKS, we can make $\bm{\Psi}$ a Gaussian random matrix to reduce the dimensionality.  Random projection can be viewed as a less-structured counterpart to classic dimensionality reduction methods like PCA and FDA. In other word, the critical information will be preserved in a less-structured way.
\subsubsection{Identity}
Particularly, the pseudo-transformation matrix $\bm{\Psi}$ can be defined as an identical matrix with ones on the main diagonal and zeros elsewhere, which indicates no dimensionality reduction is performed in the RHKS. The method is the most simple way for dimensionality reduction in KCRC, but it is effective at most time, especially in KCRC-LCD. LCD is usually constructed in relatively small size compared to the training sets, so we do not always need to perform dimensionality reduction in kernel space whose dimension is equal to the dictionary size. Thus, in the classification experiments on public database, we simply use identity matrix as $\bm{\Psi}$.
\subsubsection{Graph}
Further, we propose a graph preserving dimensionality reduction method for the pseudo-transformation matrix $\bm{\Psi}$. In the light of \cite{belkin2003laplacian}, we first construct a weighted graph with $n$ nodes ($n$ is the dictionary size, one node represents one atom in the dictionary). Then we put an edge between node $i$ and $j$ if they are close enough. Usually, there are two methods to find the nodes that we use to construct the graph. The first is $\epsilon$-neighborhoods in which node $i$ and node $j$ are connected by an edge if $\| \bm{d}_i-\bm{d}_j \|^2<\epsilon$. The second is $n$-nearest neighbors in which node $i$ and $j$ connected by an edge if $\bm{d}_i$ is among $n$ nearest neighbors of $\bm{d}_j$ or $\bm{d}_j$ is among $n$ nearest neighbors of $\bm{d}_i$. After constructing the weighted graph which contains the similarity information among atoms, we choose a measure for the weight. In \cite{belkin2003laplacian}, the following weight measure between two connected nodes is formulated as
\begin{equation}\label{egn:weight}
W_{ij}=\exp\bigg(\dfrac{\| \bm{d}_i-\bm{d}_j \|^2}{t}\bigg)
\end{equation}
besides which, there is another simple weighting method that $W_{ij}=1$ if and only if vertices $i$ and $j$ are connected by an edge. In order to group the connected nodes and separate the distant nodes as much as possible, the object function is defined as
\begin{equation}\label{egn:obj4eigenval}
\sum_{ij}(\bm{g}_i-\bm{g}_j)^2W_{ij}=2\bm{g}^T\bm{L}\bm{g}
\end{equation}
where $\bm{g}_i\ (0\leq i\leq n)$ is the map from the graph to the real sample and $\bm{L}$ is the Laplacian matrix satisfying $\bm{L}=\bm{D}-\bm{W}$ in which $\bm{D}_{ii}=\sum_jW_{ij}$. Laplacian matrix is a symmetric, positive semi-definite matrix which can be thought of as an operator on functions defined on vertices of the graph. Then we can formulate the minimization problem as
\begin{equation}\label{egn:lapmin}
\begin{aligned}
&\ \ \ \arg\min_{\bm{g}} \bm{g}^T\bm{L}\bm{g}\\
&\text{\textnormal{subj. to}}\ \ \bm{g}^T\bm{D}\bm{g}=\bm{1}
\end{aligned}
\end{equation}
which is equivalent to the solution of the following generalized eigenvalue decomposition problem:
\begin{equation}\label{egn:lapeigen}
\bm{L}\bm{g}=\lambda \bm{D}\bm{g}
\end{equation}
which is similar to the optimization problem in PCA. We let $\bm{g}_0,\bm{g}_1,\cdots,\bm{g}_{n-1}$ be the solutions of Eq.\ref{egn:lapeigen}, sorted according to their eigenvalues with $\bm{g}_0$ having the smallest eigenvalue (actually it is zero). After performing normalization on $\{\bm{g}_1,\bm{g}_2,\cdots,\bm{g}_c\}$, we take them as the pseudo-transformation matrix $\bm{\Psi}$, namely
\begin{equation}\label{eigenPsi}
\bm{\Psi}=\{\bm{g}_1,\bm{g}_2,\cdots,\bm{g}_c\}.
\end{equation}
The motivation of this dimensionality approach is quite intuitive. We construct $\bm{\Psi}$ with the graph constraint in order to combine the graph information, or to be more accurate, the similarity relation among atoms into the kernel space (after dimensionality reduction).
\subsubsection{Further Discussion}
We present four methods to perform the dimensionality reduction in the kernel space. Reducing the dimensionality in the kernel space brings several gains such as lowering the computational cost and enhancing the discrimination power.
There also exist a number of other ways to perform the dimensionality reduction in the kernel space, namely construct the matrix $\bm{\Psi}$. Empirically, if the rank of matrix $\bm{\Psi}$ stays unchanged, then different construction of $\bm{\Psi}$ will not lead to dramatical difference in classification accuracy. Thus, the matrix $\bm{\Psi}$ is not very crucial to the classifier, which is supported by the experiments conducted in \cite{zhang2012kernel}. Instead, the rank of the matrix $\bm{\Psi}$ plays a crucial part in classification accuracy. This is why even using random matrix as $\bm{\Psi}$ still serves our classifier well. In Section \uppercase\expandafter{\romannumeral4}, we conduct relevant experiments to study what the selection of the matrix $\bm{\Psi}$ will do to the classification accuracy.
\subsection{Practical KCRC Algorithms}
There are two algorithms designed for KCRC. For normal situations, $p,q$ are both set as 2. The regularized least square algorithm is adopted to solve the model with $\thickmuskip=0mu p,q=2$. Specifically, we derive the new dictionary $\thickmuskip=0mu \bm{D}'=\bm{\Psi}^T\bm{G}$ and define $\bm{P}'$ as the coding basis in kernel CRC-RLS (KCRC-RLS). Namely
\begin{equation}\label{egn:Pnew}
\bm{P}'=\big((\bm{\Psi}^T\bm{G})^T(\bm{\Psi}^T\bm{G})+\mu\cdot\bm{I}\big)^{-1}\big(\bm{\Psi}^T\bm{G}\big)^T
\end{equation}
where $\mu$ is a small constant. The query sample is transformed to $\bm{\Psi}^T\bm{K}(\bm{D},\bm{y})$. Apparently, $\bm{P}'$ is independent of $\bm{y}'$ so it can be pre-calculated. When a query $\bm{y}$ comes, the query is first transformed to the kernel space via $\bm{y'}=\bm{\Psi}^T\bm{K}(\bm{D},\bm{y})$ and then can be simply projected onto the coding basis $\bm{P}'$ via $\bm{P}'\bm{y'}$. In the decision making stage, class-specified representation residual $\| \bm{y}'- \bm{D}'_i  \hat{\bm{x}}_i \|_2$ is used for classification. Further, a $l_2$ norm term $\|\hat{\bm{x}}_i \|_2$ is added for more discriminative classification. The specific algorithm of KCRC-RLS is shown in Algorithm 1.
\par
For high level corruption and occlusion, kernel robust CRC (KRCRC) algorithm ($\thickmuskip=0mu p=2,q=1$) can be applied. Note that, $\thickmuskip=0mu \bm{D}''=\bm{\Psi}^T\bm{G}$ and $\bm{P}''_k$ are designed as the new dictionary and coding basis in kernel space respectively.
\begin{equation}\label{egn:Pnew2}
\bm{P}''=\big((\bm{\Psi}^T\bm{G})^T(\bm{\Psi}^T\bm{G})+2\mu/\sigma_k\cdot\bm{I}\big)^{-1}\big(\bm{\Psi}^T\bm{G}\big)^T
\end{equation}
\par
where $\mu,\sigma_k$ are small constants. The augmented Lagrangian function used for the optimization in Eq. \eqref{egn:KRCRC} is formulated as
\begin{equation}\label{ALMfunc}
L_{\sigma}(\bm{e},\bm{x},\bm{z})=\|\bm{e}\|_1+\mu\|\bm{x}\|_2^2+\langle \bm{z},\bm{y}''-\bm{D}''\bm{x}-\bm{e}\rangle+\dfrac{\sigma}{2}\|\bm{y}''-\bm{D}''\bm{x}\|_2^2
\end{equation}
where $\sigma$ is a positive constant that is the penalty for large representation error, and $\bm{z}$ is a vector of Lagrange multiplier. The ALM method iteratively estimates $\bm{e},\bm{x}$ for the Lagrange multiplier $\bm{z}$ via the following minimization:
\begin{equation}\label{ALMmin}
(\bm{e}_{k+1},\bm{x}_{k+1})=\arg\min_{\bm{e},\bm{x}}L_{\sigma_k}(\bm{e},\bm{x},\bm{z}_k)
\end{equation}
where $\bm{z}_{k+1}=\bm{z}_k+\sigma_k(\bm{y}''-\bm{D}''\bm{x}-\bm{e})$. According to \cite{zhang2012CRC, bertsekas1999nonlinear}, this iteration will converge to an optimal solution for Eq. \eqref{egn:KRCRC} if $\{\sigma_k\}$ is a monotonically increasing sequence.
\par
The minimization process in Eq. \eqref{ALMmin} can be implemented by optimizing $\bm{e},\bm{x}$ alternatively and iteratively:
\begin{equation}\label{egn:ALMminnew}
\begin{aligned}
&\ \bm{x}_{k+1}=\arg\min_{\bm{x}}L_{\sigma_k}(\bm{x},\bm{e}_k,\bm{z}_k),\\
&\bm{e}_{k+1}=\arg\min_{\bm{e}}L_{\sigma_k}(\bm{x}_{k+1},\bm{e},\bm{z}_k),
\end{aligned}
\end{equation}
which has the closed-form solution as follows:
\begin{equation}\label{egn:ALMminSol}
\begin{aligned}
&\ \bm{x}_{k+1}=(\bm{D}''^T\bm{D}''+2\mu/\sigma_k\bm{I})^{-1}\bm{D}''^T(\bm{y}''-\bm{e}_k+\bm{z}_k/\sigma_k))\\
&\ \ \ \ \ \ \ \ =\bm{P}''_k(\bm{y}''-\bm{e}_k+\bm{z}_k/\sigma_k),\\
&\ \bm{e}_{k+1}=S_{1/\sigma_k}(\bm{y}''-\bm{D}''\bm{x}_{k+1} + \bm{z}_k/\sigma_k),
\end{aligned}
\end{equation}
where the function $S_{\alpha},\alpha\geq0$ is the soft-thresholding (shrinkage) operator given by
\begin{equation}\label{ShOper}
 S_{\alpha}(h) = \left\{ {\begin{array}{*{20}{l}}
{h-\alpha,\ \ \ \text{\textnormal{if}}\ x\ge\alpha}\\
{h+\alpha,\ \ \ \text{\textnormal{if}}\ x\le\alpha}\\
{0,\;\ \ \ otherwise}
\end{array}} \right..
\end{equation}
If $\bm{h}$ represents a $n$-dimensional vector, then $S_{\alpha}(\bm{h})$ is given by $\{S_{\alpha}(\bm{h}_1),S_{\alpha}(\bm{h}_2),\cdots,S_{\alpha}(\bm{h}_n) \}$. Similar to the KCRC-RLS, the coding basis $\bm{P}''_k$ is independent of $\bm{y}''$ for the given $\sigma_k$, so the set of projection matrices $\{\bm{P}_k\}$ can also be pre-calculated. Once a query sample $\bm{y}$ comes, it is first transformed in the kernel space via $\bm{\Psi}^T\bm{K}(\bm{D},\bm{y})$ and then projected onto $\bm{P}''_k$ via $\bm{P}''_k\bm{y}''$. After performing the iterative minimization above, a classification strategy similar to KCRC-RLS is applied in KRCRC. Details of KRCRC is given in Algorithm 2.
\par

\begin{center}
\begin{tabular}{l @{}}\hline
\multicolumn{1}{c}{Algorithm 1: KCRC-RLS} \\\hline
1. Normalize the columns of $\thickmuskip=0mu \bm{D}'=\bm{\Psi}^T\bm{G}$ to unit $l_2$-norm.\\
2. Represent $\bm{y}'=\bm{\Psi}^T\bm{K}(\bm{D},\bm{y})$ over dictionary $\bm{D}'$ by\\
\multicolumn{1}{c}{$\hat{\bm{x}}=\bm{P}'\bm{y}'$} \\
\ \ \ \ where $\bm{P}'=(\bm{D}'^T\bm{D}'+\mu\bm{I})^{-1}\bm{D}'^T$.\\
3. Obtain the regularized residuals\\
\multicolumn{1}{c}{$r_i=\| \bm{y}'- \bm{D}'_i  \hat{\bm{x}}_i\|_2/\| \hat{\bm{x}}_i \|_2$} \\
\ \ \ \ where $\hat{\bm{x}}_i$ is the coding coefficients associated with class\\
\ \ \ \ $i$ over $\bm{P}'$.\\
4. Output the identity of $\bm{y}'$ (class label) as\\
\multicolumn{1}{c}{$identity(\bm{y}')=\arg\min_i(r_i)$.} \\
\hline
\end{tabular}
\end{center}
\
\begin{center}
\begin{tabular}{l @{}}\hline
\multicolumn{1}{c}{Algorithm 2: KRCRC} \\\hline
1. Normalize the columns of $\thickmuskip=0.5mu \bm{D}''=\bm{\Psi}^T\bm{G}$ to unit $l_2$-norm.\\
2. Input $\thickmuskip=0mu\bm{y}''=\bm{\Psi}^T\bm{K}(\bm{D},\bm{y})$, $\bm{x}_0$, $\bm{e}_0$, $\thickmuskip=0mu k=1$ and $\tau>0$.\\
3. Proceed if $|\bm{x}_{k+1}-\bm{x}_k|_2 > \tau$ is true. If not, output $\hat{\bm{e}},\hat{\bm{x}}$\\
\ \ \ \ and go to step 5.\\
4. Do the following iteration:\\
\multicolumn{1}{c}{$\bm{x}_{k+1}=\bm{P}''_k(\bm{y}''-\bm{e}_k+\bm{z}_k/\sigma_k)$} \\
\multicolumn{1}{c}{$\bm{e}_{k+1}=S_{1/\sigma_k}(\bm{y}''-\bm{D}''\bm{x}_{k+1} + \bm{z}_k/\sigma_k)$} \\
\multicolumn{1}{c}{$\bm{z}_{k+1}=\bm{z}_k+\sigma_k(\bm{y}''-\bm{D}''\bm{x} _{k+1}-\bm{e}_{k+1})$} \\
\ \ \ \ where $\thickmuskip=0mu \bm{P}''_k=(\bm{D}''^T\bm{D}''+2\mu/\sigma_k\bm{I})^{-1} \bm{D}''^T$ and $\thickmuskip=0mu S_{\alpha},\alpha\geq0$\\
\ \ \ \ is the shrinkage coefficient. $\thickmuskip=0mu k\leftarrow k+1$ and go to step 3.\\
5. Represent $\bm{y}''$ over dictionary $\bm{D}''$ by the converged $\bm{x}$.\\
6. Obtain the regularized residuals\\
\multicolumn{1}{c}{$r_i=\| \bm{y}''- \bm{D}''_i  \hat{\bm{x}}_i\|_2/\| \hat{\bm{x}}_i \|_2$} \\
\ \ \ \ where $\hat{\bm{x}}_i$ is the coding coefficients related to class $i$.\\
7. Output the identity of $\bm{y}''$ (class label) as\\
\multicolumn{1}{c}{$identity(\bm{y}'')=\arg\min_i(r_i)$.} \\
\hline
\end{tabular}
\end{center}
\section{On the Locality Constrained Dictionary}
This section elaborates the locality constrained dictionary. Additionally, we present some typical distances used in LCD for similarity measurement and further introduce a distance fusion model, followed by the introduction of the KCRC method combined with LCD, termed as KCRC-LCD.
\subsection{Locality Constrained Dictionary}
Most collaborative representation based methods \cite{zhang2012CRC,zhu2012multi,waqas2013collaborative} employ all the high-dimensional training samples as the global dictionary. They may work fine when the global dictionary is small, but the classification becomes intractable with increasingly more training samples. To tackle with this problem, we propose the LCD that utilizes the K-NN classifier to measure the similarities between the query sample and all atoms in the global dictionary, and then selects $K$ nearest atoms as the local dictionary. The locality in LCD ensures discrimination, efficiency and robustness of KCRC. Compared to the locality constrained dictionary proposed in \cite{zhou2013locality}, we adopt a more straightforward way to constrain the locality, which needs no learning and training process, greatly reducing the computational cost in training. Under such locality constrained dictionary, scaling to a large number of categories dose not require adding new features, because the discriminative distance function need only be defined for similar enough samples. From biological and psychological perspective, similarity between samples is the most important criteria to recognize and classify objects for human brains. So intuitively speaking, the proposed locality constrained dictionary with various optional discriminative distances makes our KCRC approach more scalable, discriminative, efficient, and most importantly, free from the curse of high-dimensional feature space. Moreover, the kernel idea within KCRC well suits the idea of locality both mathematically and experimentally.
\par
Define $Dist(\bm{d}',\bm{d}'')$ as the distance metric between atom $\bm{d}'$ and $\bm{d}''$. To be simple, we adopt the $l_2$ distance as example in formulation, namely $\thickmuskip=2mu Dist(\bm{d}',\bm{d}'')=\|\bm{d}'-\bm{d}''\|_2$. We need to calculate the distance between every atom $\bm{d}_{[k]}$ and the query sample $\bm{y}$ first. Then the LCD can be obtained via the following optimization:
\begin{equation}\label{egn:LCD}
\begin{aligned}
&\ \ \ \arg \min_{\{t_1,t_2,\cdots,t_K\}} \sum_{m=1}^{K} Dist(\bm{d}_{[t_m]},\bm{y})\\
&\text{\textnormal{subj.\ to}} \ \ 1\leq t_i \not= t_j \leq n,\ for\ \forall i \not= j
\end{aligned}
\end{equation}
where $\bm{d}_{[t_1]},\bm{d}_{[t_2]},\cdots,\bm{d}_{[t_K]}$ denote different atoms in the global dictionary. In fact, to solve Eq. \eqref{egn:LCD} is to find the $K$ atoms that are located nearest to the query sample. As a result, the LCD is obtained as $\thickmuskip=0mu \bm{D}_{lc}=\{\bm{d}_{[t_1]},\bm{d}_{[t_2]},\cdots,\bm{d}_{[t_K]}\}$. Moreover, the computational complexity of solving Eq. \eqref{egn:LCD} is $O(n\log n)$, which is efficient enough to perform in large-scale image databases. Note that, when $\thickmuskip=0mu K=n$, KCRC-GD becomes a special case of KCRC-LCD.
\subsection{Discriminative Distances for Similarity Measure}
In the previous subsection, we simply use the $l_2$ distance as an example. In fact, there are many discriminative distances for similarity measurement. Several well-performing distances are introduced in \cite{zhang2006svm}, i.e., Mahalanobis distance, $\chi^2$ distance \cite{leung2001representing}, marginal distance \cite{levina2002statistical}, tangent distance \cite{simard2012transformation}, shape context based distance \cite{belongie2002shape} and geometric blur based distance \cite{berg2001geometric}, etc. Each can be used to measure the similarity in order to construct a well-performing LCD. These distances can either be used alone or used in conjunction with each other, making the LCD flexible and adaptive. We will review some of the discriminative distances in this subsection.
\subsubsection{General Pixel Similarity Measure}
We consider several classical pixel distance metrics below. Euclidean distance ($l_2$ distancep) is the most popular similarity measure. It is simple yet effective in certain situations and defined as
\begin{equation}\label{euclidean}
Dist(\bm{d}',\bm{d}'')=\|\bm{d}'-\bm{d}''\|_2.
\end{equation}
\par
City block distance, also known as Manhattan distance, assumes that it is only possible to travel along pixel grid lines from one pixel to another. This distance metric is defined as
\begin{equation}\label{cityblock}
Dist(\bm{d}',\bm{d}'')=\|\bm{d}'-\bm{d}''\|_1.
\end{equation}
\par
Chessboard distance metric assumes that you can make moves on the pixel grid as if you were a King making moves in chess, i.e. a diagonal move counts the same as a horizontal move. The metric is given by:
\begin{equation}\label{chess}
Dist(\bm{d}',\bm{d}'')=\|\bm{d}'-\bm{d}''\|_{\infty}.
\end{equation}
There are a lot of other general pixel distances that can be utilized in our framework, such as correlation distance, Mahalanobis distance, etc.
\subsubsection{Texture Similarity Measure}
We present some texture similarity measure as examples below. The $\chi^2$ distance is proposed in \cite{leung2001representing} for texture similarity measure. The main idea of the $\chi^2$ distance is to construct a vocabulary of 3D textons by clustering a set of samples. Associated with each texton is an appearance vector which characterizes the local irradiance distribution. The similarity can be measured by characterizing samples with these 3D textons.
\par
In the view of statistics, marginal distance \cite{levina2002statistical} is another version of the $\chi^2$ distance. They both measure the difference between two joint distribution of texture response. The difference is that marginal distance metric simply sums up the distance between response histograms from each filter while the $\chi^2$ distance metric measures the similarity of the two joint distribution by comparing the histogram of textons.
\par
As another texture similarity measure initially used for hand-written digits recognition, Tangent distance \cite{simard2012transformation} is defined to compute the minimum distance between the linear surfaces that best approximate the non-linear manifolds of different sample categories. These linear surfaces, which are crucial to Tangent distance, are derived from the images by including perturbations from small affine transformation of the spatial domain and change in the thickness of pen-stroke.
\subsubsection{Shape Similarity Measure}
The shape in an image can be represented by a set of points, with a descriptor at a fixed point to measure the relative position to that point. These descriptors are iteratively matched using a deformation model. Shape context based distance \cite{belongie2002shape} is derived from the discrepancy left in the final matched shapes and a score that denotes how far the deformation is from an affine transformation. Various shape descriptors can be defined on a gray scale image, for example, the shape context descriptor on the edge map \cite{mori2002estimating}, the SIFT descriptor \cite{lowe2004distinctive}, and the geometric blur descriptor \cite{berg2001geometric}, optimized local shape descriptor \cite{taati2011local}, etc.
\subsubsection{Unified Similarity Measure}
\begin{figure}[h]
  \renewcommand{\captionlabelfont}{\footnotesize}
  \setlength{\abovecaptionskip}{0pt}
  \setlength{\belowcaptionskip}{0pt}
  \centering
  \includegraphics[width=3in]{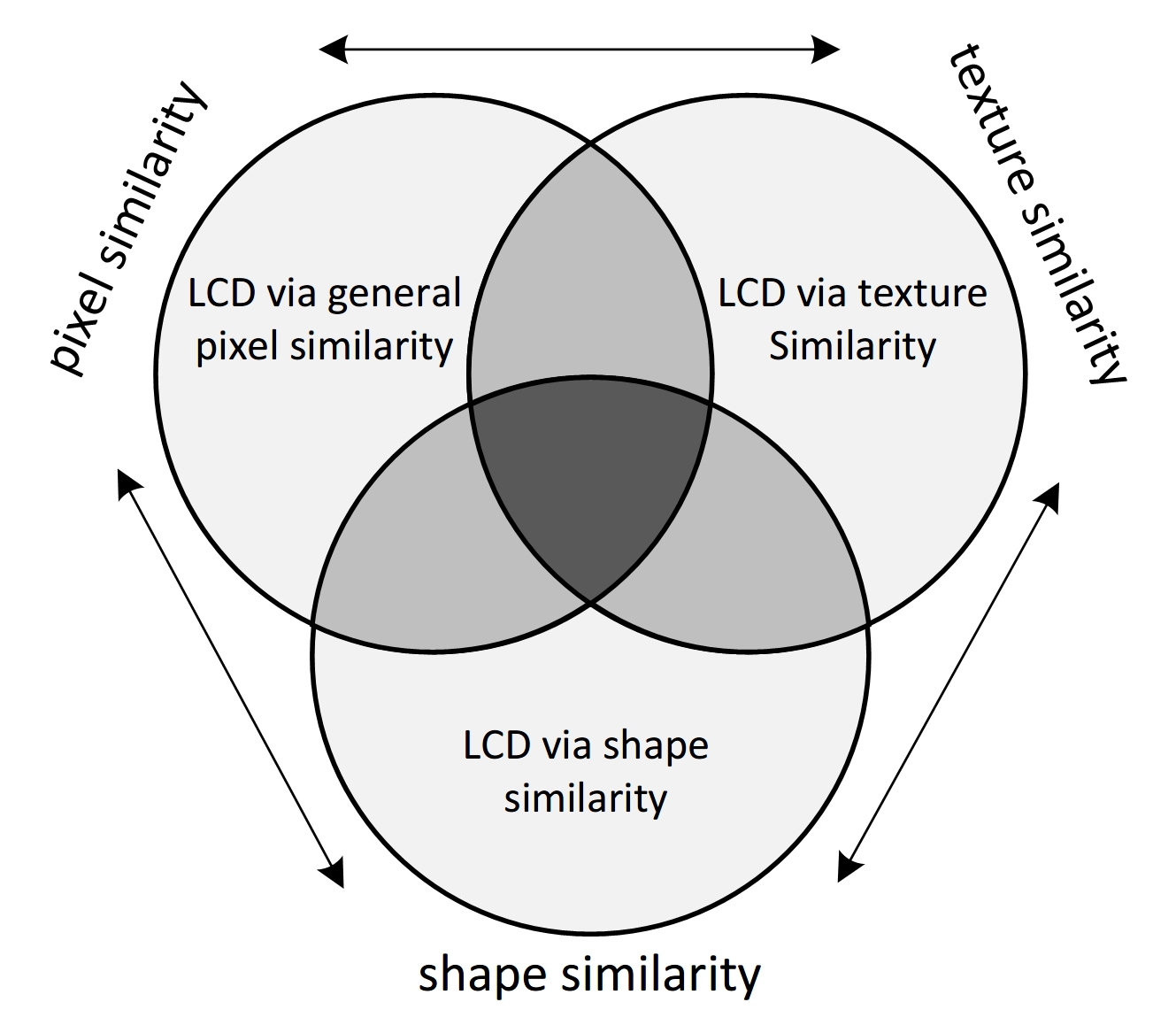}\\
  \caption{.\ \footnotesize Venn diagram of LCDs constructed via general pixel similarity, texture similarity and shape similarity.}\label{fig:fig1}
\end{figure}

We use a simple and intuitive method to combine general pixel similarity, texture similarity and shape similarity into a unified locality constrained dictionary. Assume that we need to construct a LCD with size $K$ out of total $N$ training samples. First, we enforce locality to dictionary via general pixel similarity, texture similarity and shape similarity, obtaining three LCD with size $K$: $\bm{D}_{lc}^{[pixel]}$, $\bm{D}_{lc}^{[texture]}$ and $\bm{D}_{lc}^{[shape]}$. From the sets perspective, these three dictionaries constructed via different similarity measure can be viewed in Venn diagram as shown in Fig. \ref{fig:fig1}. Specifically, an atom in dictionary represents an element in a set, so a LCD can be regarded as a set with $K$ elements. According to the demand of the given task, we can use different combinations of similarity measures to construct the LCD. Mathematically, we achieve the combination of similarity measures by getting the union of the corresponding LCDs, i.e., $\bm{D}_{lc}=\bm{D}_{lc}^{[texture]}\cup \bm{D}_{lc}^{[shape]}$ for the combination of texture and shape similarity measure. With unified similarity measure, the distance metric used in the kernel function becomes a linear combination of the distances that are utilized to construct the new LCD. Normally, we suppose the weight of each distance metric in the unified distance is equal. However, to use unified similarity measure could add to computational cost, so we do not recommend to use it under normal circumstance. Note that, the same type similarity measures can also be unified by the proposed method with similar procedure.
\subsection{KCRC-LCD Algorithm}
Intuitively, we use the locality constrained dictionary $\bm{D}_{lc}$ in place of global dictionary $\bm{D}$ and then perform the KCRC algorithm on $\bm{D}_{lc}$. The KCRC-LCD algorithm is as follows:
\begin{center}
\begin{tabular}{l @{}}\hline
\multicolumn{1}{c}{Algorithm 3: Naive KCRC-LCD} \\\hline
1. Compute the distances between the query sample and\\
\ \ \ \ all training samples, and pick the nearest $K$ neighbors.\\
2. If the $K$ neighbors have the same labels, the query is\\
\ \ \ \ labeled and exit; Else, construct the LCD $\bm{D}_{lc}$ with the\\
\ \ \ \ $K$ labeled neighbors and goto Step 3.\\
3. Convert the pairwise distance into a kernel matrix via\\
\ \ \ \ kernel trick and utilize KCRC approach with dictionary\\
\ \ \ \ $\bm{D}_{lc}$ instead of the global dictionary $\bm{D}$.\\
4. Output the label of the query sample.\\
\hline
\end{tabular}
\end{center}
\par
The naive version of KCRC-LCD performs slowly because it has to compute the distances of the query to all training samples. Inspired by \cite{zhang2006svm}, we consider to boost the efficiency in the coarse-to-fine framework which is similar to the human perception procedure that human first perform a fast coarse pruning and then recognize the object by details. The practical version of KCRC-LCD is as follows:
\begin{center}
\begin{tabular}{l @{}}\hline
\multicolumn{1}{c}{Algorithm 4: Pratical KCRC-LCD} \\\hline
1. Compute the coarse distances (i.e. Euclidean distance)\\
\ \ \ \ between the query sample and all training samples, and\\
\ \ \ \ pick the nearest $K_c$ neighbors. ($K_c\geq K_f$)\\
2. Compute the fine distances between the query sample\\
\ \ \ \ and pick the nearest $K_f$ neightbors.\\
2. If the $K_f$ neighbors have the same labels, the query is\\
\ \ \ \ labeled and exit; Else, construct the LCD $\bm{D}_{lc}$ with the\\
\ \ \ \ $K_f$ labeled neighbors and goto Step 3.\\
3. Convert the pairwise distance into a kernel matrix via\\
\ \ \ \ kernel trick and utilize KCRC approach with dictionary\\
\ \ \ \ $\bm{D}_{lc}$ instead of the global dictionary $\bm{D}$.\\
4. Output the label of the query sample.\\
\hline
\end{tabular}
\end{center}
\par
\section{Experiments and Results}
\subsection{Evaluation of Dimensionality Reduction in Kernel Space}
\begin{figure*}[!htb]
  \renewcommand{\captionlabelfont}{\footnotesize}
  \centering
  \includegraphics[width=6.1in]{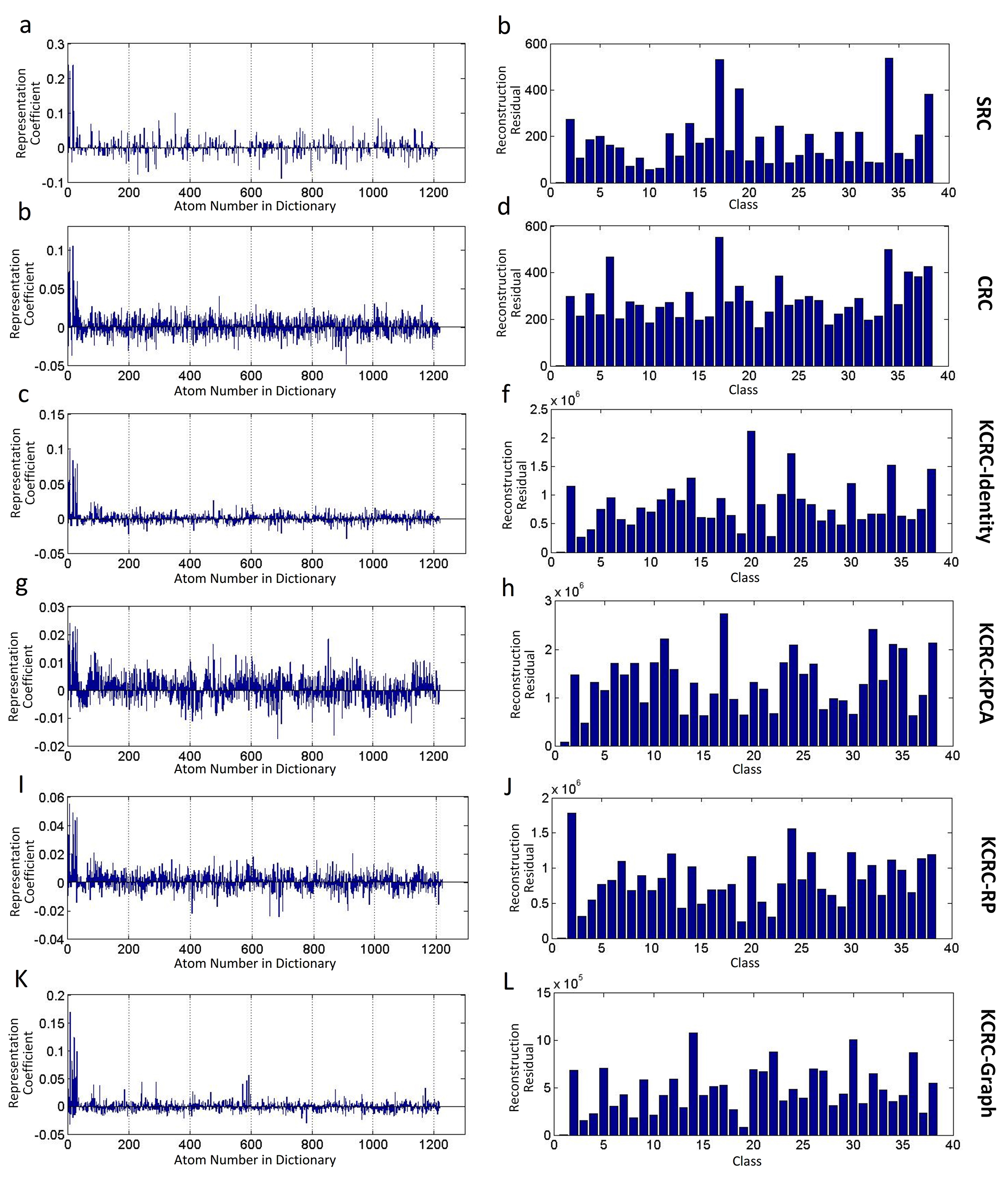}\\
  \caption{.\ \footnotesize (a) Representation coefficients obtained by SRC. (b) Reconstruction Residuals obtained by SRC. (c) Representation coefficients obtained by CRC. (d) Reconstruction Residuals obtained by CRC. (e) Representation coefficients obtained by KCRC-Identity. (f) Reconstruction Residuals obtained by KCRC-Identity. (g) Representation coefficients obtained by KCRC-KPCA. (h) Reconstruction Residuals obtained by KCRC-KPCA. (i) Representation coefficients obtained by KCRC-RP. (j) Reconstruction Residuals obtained by KCRC-RP. (k) Representation coefficients obtained by KCRC-Graph. (l) Reconstruction Residuals obtained by KCRC-Graph.}\label{fig:DR}
\end{figure*}
We evaluate the dimensionality reduction from two aspects. First we compare the representation coefficients and reconstruction residuals that are obtained by SRC, CRC, KCRC with no dimensionality reduction (KCRC-Identity), KCRC with KPCA dimensionality reduction (KCRC-KPCA), KCRC with random projection (KCRC-RP) and KCRC with graph projection (KCRC-Graph). Second, we compare the recognition accuracy of these methods in extended Yale B face database.
\par
We randomly select 38 images per person (32 person, totally 1216 images) in extended Yale B database \cite{georghiades2001few} as training samples. For the computation of representation coefficients and reconstruction residuals, we use a single fixed test sample for better comparison. To simplify the experiment, we only use the global dictionary since the experiment focuses on the dimensionality reduction methods in kernel space. In the recognition accuracy test, we follow the same experiments settings as \cite{wright2009robust} by randomly selecting 38 images per person (32 person, totally 1216 images) and using the remaining images as test samples. The results are averaged over 20 times experiments.
\par
Fig. \ref{fig:DR} gives the representation coefficients and reconstruction residuals of SRC, CRC, KCRC-Identity, KCRC-KPCA, KCRC-RP  and KCRC-Graph. We can see all of these five approaches tell the correct label (the first class) of the test sample. It can be obtain from Fig. \ref{fig:DR} that KCRC-Identity and KCRC-Graph achieve better sparsity, similar to the representation coefficients of SRC. Moreover, the reconstruction residuals of all these approaches indicate the first class has the fewest reconstruction residual. Table. \ref{tab:DR} shows the recognition accuracy of SRC, CRC, KCRC-Identity, KCRC-KPCA, KCRC-RP and KCRC-Graph on extended Yale B database. We can see CRC has the best recognition accuracy of these five methods. However, all these approaches are of the same level discrimination ability since the difference between the highest recognition rate and the lowest is less than 1\%.
\begin{table}[htb]
  \renewcommand{\captionlabelfont}{\footnotesize}
  \setlength{\abovecaptionskip}{2pt}
  \setlength{\belowcaptionskip}{2pt}
  \centering
  \caption{.\ \footnotesize Recognition results on extended Yale B database. 504 random projection features and the global dictionary (1216 atoms) are adopted. The results below are averaged over 10 times experiments.}\label{tab:DR}
\begin{tabular}{c c}\hline
Method & Accuracy(\%)\\\hline
SRC & 97.15 \\
CRC & 97.67 \\
KCRC-Identity & 97.23 \\
KCRC-KPCA & 97.07 \\\
KCRC-RP & 96.61 \\
KCRC-Graph & 97.35 \\\hline
\end{tabular}
\end{table}

\subsection{Experiments on Data with the Same Direction Distribution}
\begin{figure*}[htb]
  \setlength{\abovecaptionskip}{0pt}
  \setlength{\belowcaptionskip}{0pt}
  \renewcommand{\captionlabelfont}{\footnotesize}
  \centering
  \includegraphics[width=5.6in]{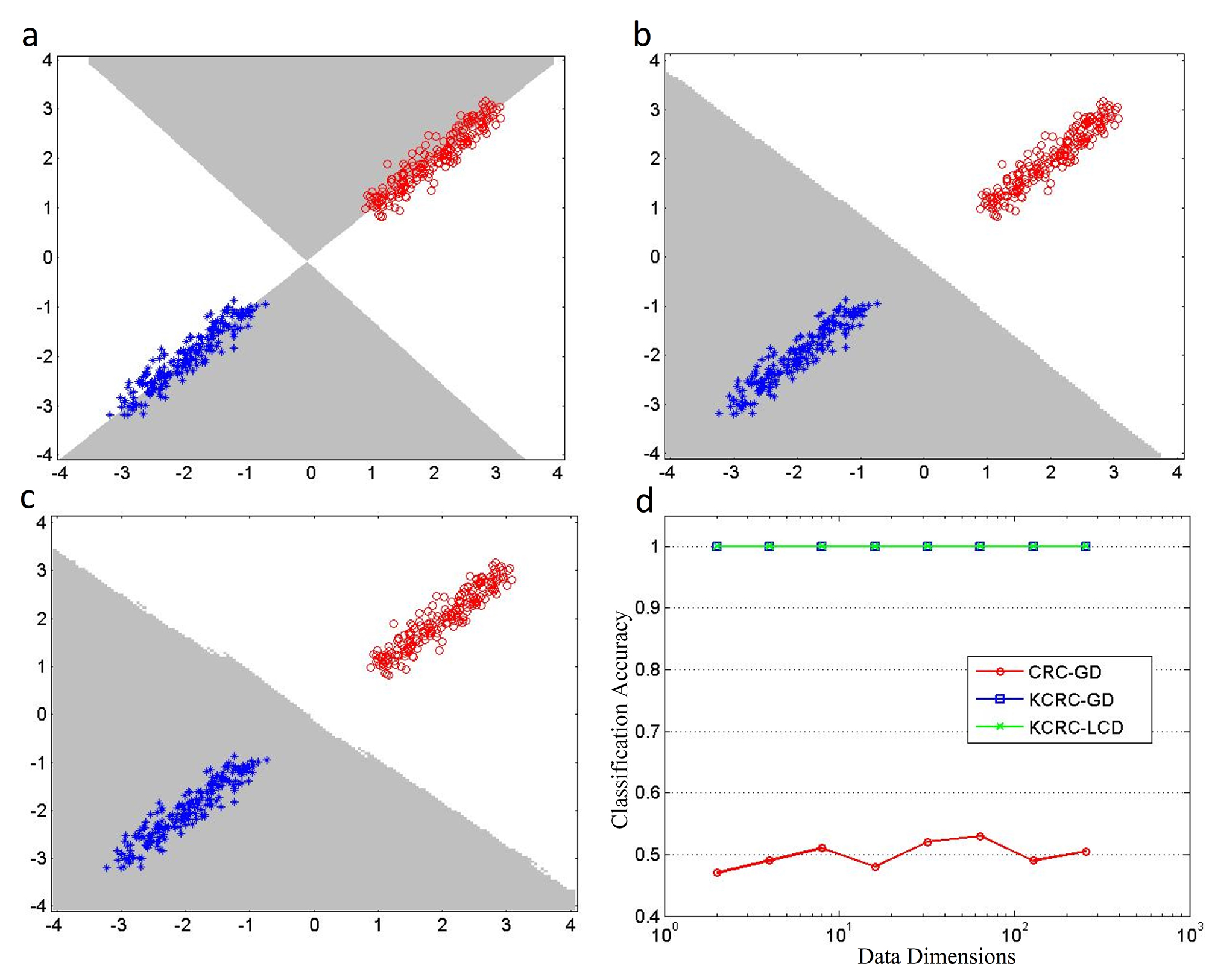}\\
  \caption{.\ \footnotesize Performance comparison on data with the same direction distribution. Test samples are from the entire surface whose predicted labels are indicated by gray or white. 2-D decision boundaries obtained by (a) CRC with global dictionary (CRC-GD), (b) KCRC with global dictionary (KCRC-GD), (c) KCRC with locality constrained dictionary (KCRC-LCD). (d) Classification accuracy vs. dimensionality. }\label{fig:fig2}
\end{figure*}
We evaluate the performance on data with the same direction distribution. In Fig. \ref{fig:fig2}, we compare 3 classifiers: CRC-GD, KCRC-GD and KCRC-LCD. Two-class training data $\bm{Q},\bm{W}$ with $m$-dimension are generated for the experiment. Each feature of $\bm{Q},\bm{W}$ uniformly takes value from the interval $[1,3]$ and $[-3,-1]$ respectively, corrupted by Gaussian noise with zero mean and 0.15 variance. Then, we let $m$ vary from 2 to 256 and perform the experiment. The results show that both KCRC-GD and KCRC-LCD can perfectly classify data with the same direction distribution while CRC performs poorly. This experiment shows both KCRC-GD and KCRC-LCD could handle data with special distribution, i.e. the same direction distribution in this case. Thus, kernel function makes our proposed approach more prepared for unknown data distribution than conventional CRC.
\subsection{Experiments on Public Databases}
This subsection evaluates our approach on public databases. Reliable results are obtained by 20 times repeated experiments with different random splits of the training and test images.
\subsubsection{MNIST}
\begin{figure*}[htb]
  \renewcommand{\captionlabelfont}{\footnotesize}
  \centering
  \includegraphics[width=6in]{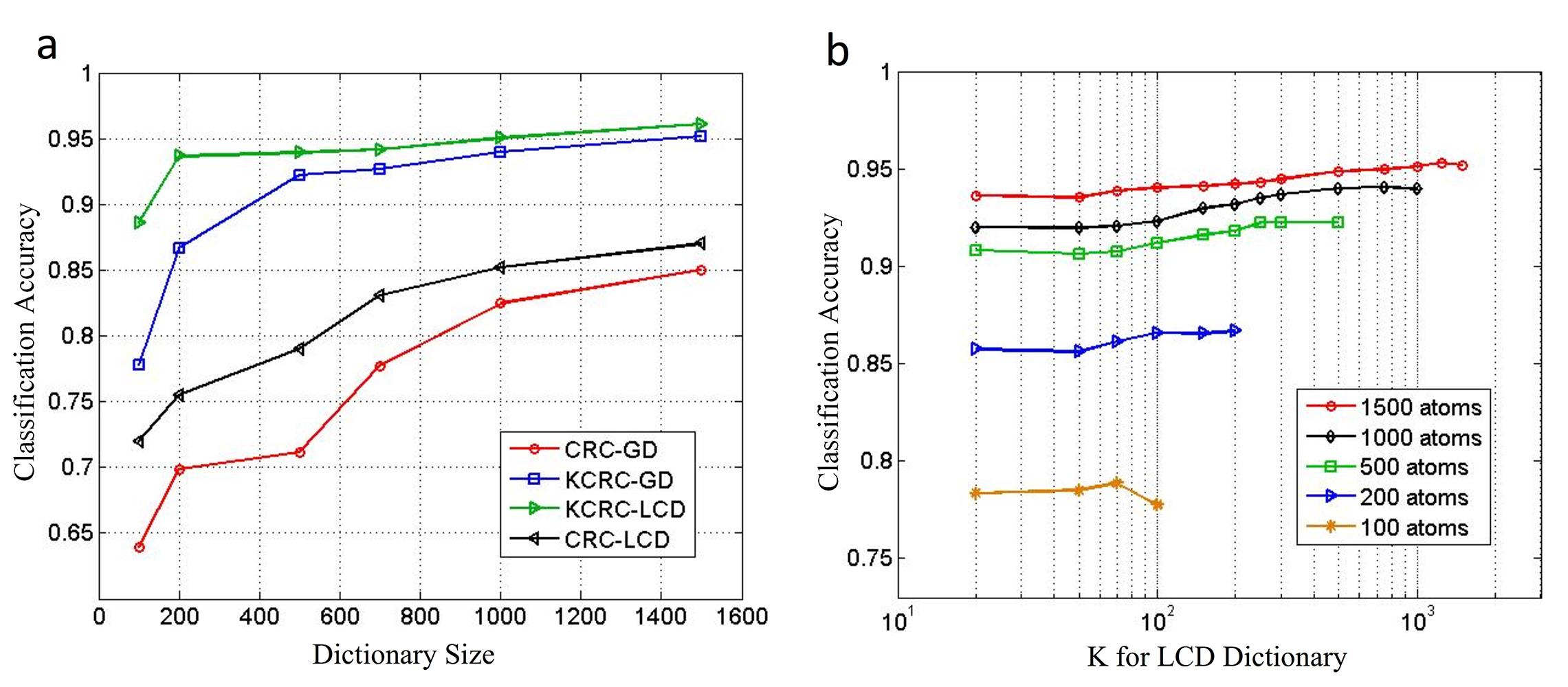}\\
  \caption{.\ \footnotesize (a) Performance comparison on MNIST under different dictionary size. (b) KCRC-LCD with different size of the global dictionary that generates the LCD under different $K$ settings. Note that, $l_2$ distance and $28\times 28$ raw pixel features are used for similarity measure and classification.}\label{fig:mnist}
\end{figure*}
The MNIST database \cite{lecun1998gradient} of handwritten digits contains 60,000 samples (10 digits) for training and 10,000 for testing. For the experimental settings, $l_2$ distance and $28\times 28$ raw pixel features are used for similarity measure and classification. We evaluate our approach via different dictionary size 100, 200, 500, 700, 1000 and 1500, namely 10, 20, 50, 70, 100 and 150 samples for training per category. For settings of KCRC-LCD and CRC-LCD, we use the global dictionary of size 2000 (200 training samples per category) to generate the LCD and set $K$ for LCD as 100, 200, 500, 700, 1000 and 1500 for comparison. Experimental results in Fig. \ref{fig:mnist}(a) show KCRC-LCD has the best performance compared to CRC-LCD, CRC-GD and KCRC-GD in the MNIST database. From Fig. \ref{fig:mnist}(b), it can be learned that $K$ has slight impact on classification accuracy if the global dictionary is fixed (atom number of GD stays unchanged).
\subsubsection{Extended Yale B Faces}
\begin{figure*}[htb]
  \renewcommand{\captionlabelfont}{\footnotesize}
  \centering
  \includegraphics[width=6in]{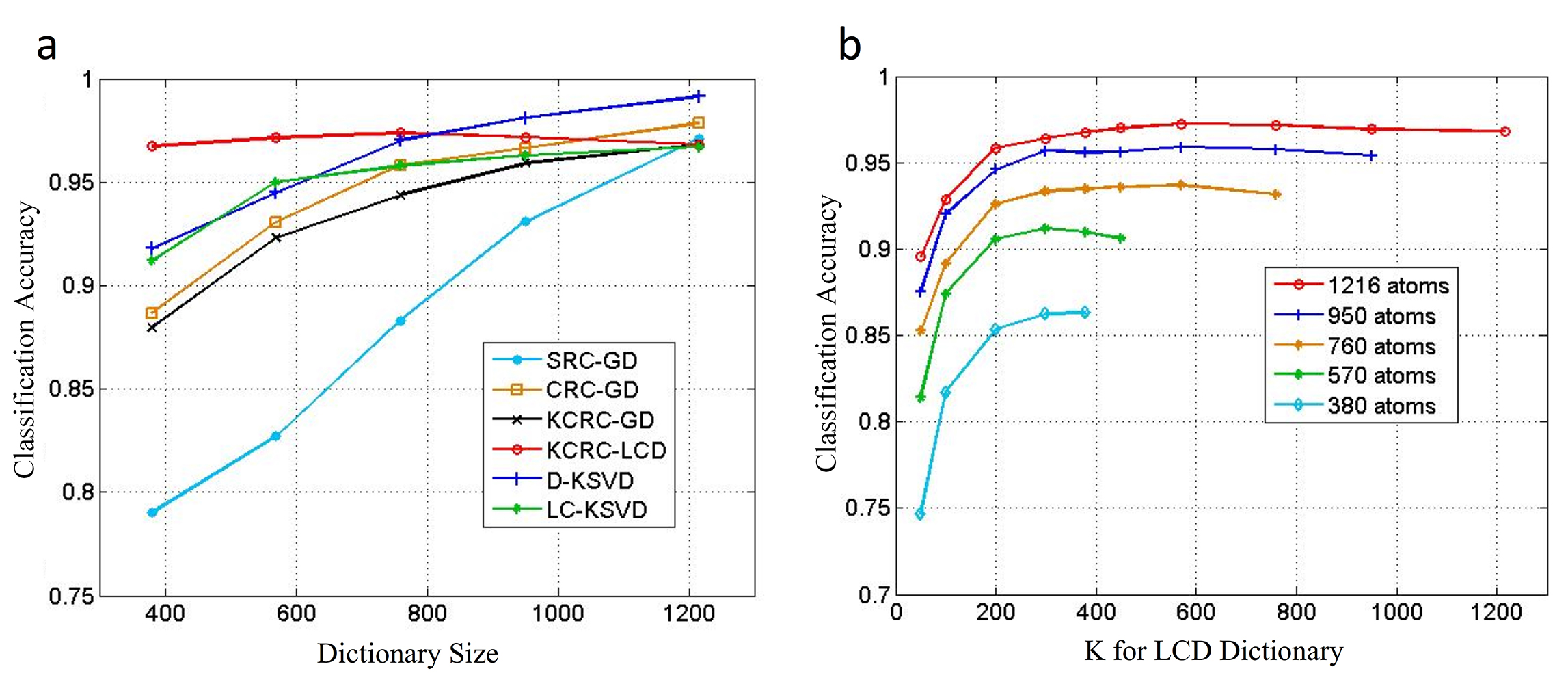}\\
  \caption{.\ \footnotesize (a) Performance comparison with SRC-GD, CRC-GD, KCRC-GD, KCRC-LCD, LC-KSVD \cite{jiang2013label,jiang2011learning} and discriminative K-SVD (D-KSVD) \cite{zhang2010discriminative} on extended Yale B database under different dictionary size. (b) KCRC-LCD with different size of the global dictionary that generates the LCD under different $K$ settings. Note that, $l_2$ distance and the random projection features are used for similarity measure and classification.}\label{fig:yaleB}
\end{figure*}
The extended Yale B database consists of 2414 frontal face images of 38 individuals \cite{georghiades2001few}. The cropped $192\times 168$ face images are taken under various lighting conditions \cite{georghiades2001few}. For each person, we randomly select 32 images for training and the remaining for testing. Therefore, there will be 1216 training images and 1198 test images. For the experimental settings, $l_2$ distance and 504-dimension random projection features \cite{wright2009robust,jiang2011learning} are used for similarity measure and classification. We evaluate our approach via different dictionary size 380, 570, 760, 950 and 1216, namely 10, 15, 20, 25 and 32 samples for training per category. For settings of KCRC-LCD, we use the global dictionary of size 1216 (32 training samples per person) to generate the LCD and set $K$ for LCD as 380, 570, 760, 950 and 1216 for comparison. Experimental results are given in Fig. \ref{fig:yaleB}. In Fig. \ref{fig:yaleB}(a), KCRC-LCD has better classification accuracy than the other approaches when dictionary size is small. It is mostly because the global dictionary we use to generate LCD is more informative than the small size dictionary, and LCD itself is designed to be adaptive to use the important information for classification. We can also learn that the classification accuracy of KCRC-LCD no longer stands out when dictionary size becomes 1216. When dictionary size comes to 1216, $K$ for LCD is also equal to 1216, making KCRC-LCD degenerate to KCRC-GD. That is to say, we do not have enough training samples to construct a discriminative LCD. Moreover, when $K$ becomes larger, the locality of LCD becomes weaker as well, leading to less discrimination power. It is obtained from Fig. \ref{fig:yaleB}(b) that LCD already has enough critical information to proceed the classification when $K$ reaches 380. Therefore, we can conclude that the performance ceiling for LCD has been reached when $K=380$ in this case. Compared to the global dictionary with 1214 training samples, LCD with only 380 atoms can achieve similar or even better classification accuracy.
\subsubsection{Caltech101}
\begin{figure*}[htb]
  \renewcommand{\captionlabelfont}{\footnotesize}
  \centering
  \includegraphics[width=6in]{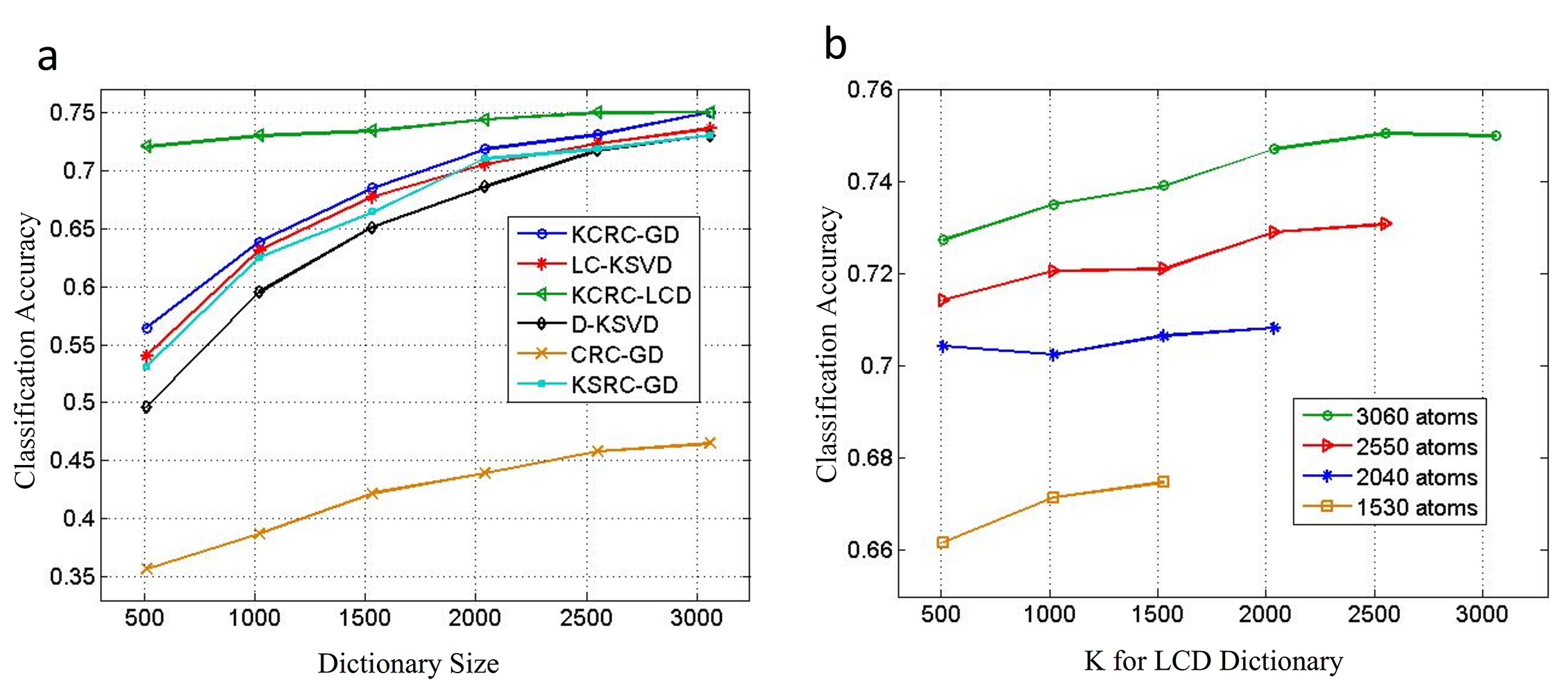}\\
  \caption{.\ \footnotesize (a) Performance comparison with KCRC-GD, KCRC-LCD, CRC-GD, LC-KSVD \cite{jiang2013label,jiang2011learning}, D-KSVD \cite{zhang2010discriminative} and KSRC-GD \cite{zhang2012kernel} on Caltech101 under different dictionary size. (b) KCRC-LCD with different size of the global dictionary that generates the LCD under different $K$ settings. Note that, $l_2$ distance and the spatial pyramid features are used for similarity measure and classification.}\label{fig:cal101}
\end{figure*}
The Caltech101 database \cite{fei2007learning} contains 9,144 images from 102 classes (101 objects and a background class). We train on 5, 10, 15, 20, 25, 30 samples per category (dictionary size is 510, 1020, 1530, 2040, 2550, 3060 respectively) and test on the rest. Note that, we use the global dictionary of size 3060 (30 training samples per category) to generate the LCD and set $K$ for LCD as 510, 1020, 1530, 2040, 2550 and 3060 for comparison. $l_2$ distance and the spatial pyramid features are used for similarity measure and classification. Results in Fig. \ref{fig:cal101}(a) show KCRC-LCD outperforms other competitive approaches in Caltech101 database, especially when dictionary size is small. As we can see in Fig. \ref{fig:cal101}(a), the classification accuracy is improved little when dictionary size is high. The reason is similar to the previous experiment on extended Yale B database. It is due to the lack of extra training samples, or in other word, extra discriminative information, to construct LCD since KCRC-LCD becomes KCRC-GD when $K$ equals to the 3060. From Fig. \ref{fig:cal101}(b), we can learn that when LCD has obtained the most discriminative and crucial atoms in the dictionary, keeping increasing $K$ will not help the classification accuracy much. In fact, if we perform the experiment in Fig. \ref{fig:cal101}(b) with smaller $K$, it will end up like the curves in Fig. \ref{fig:yaleB}(b) where $K$ obviously has a saturation point for classification accuracy.
\subsubsection{Caltech256}
\begin{figure}[htb]
  \renewcommand{\captionlabelfont}{\footnotesize}
  \centering
  \includegraphics[width=3in]{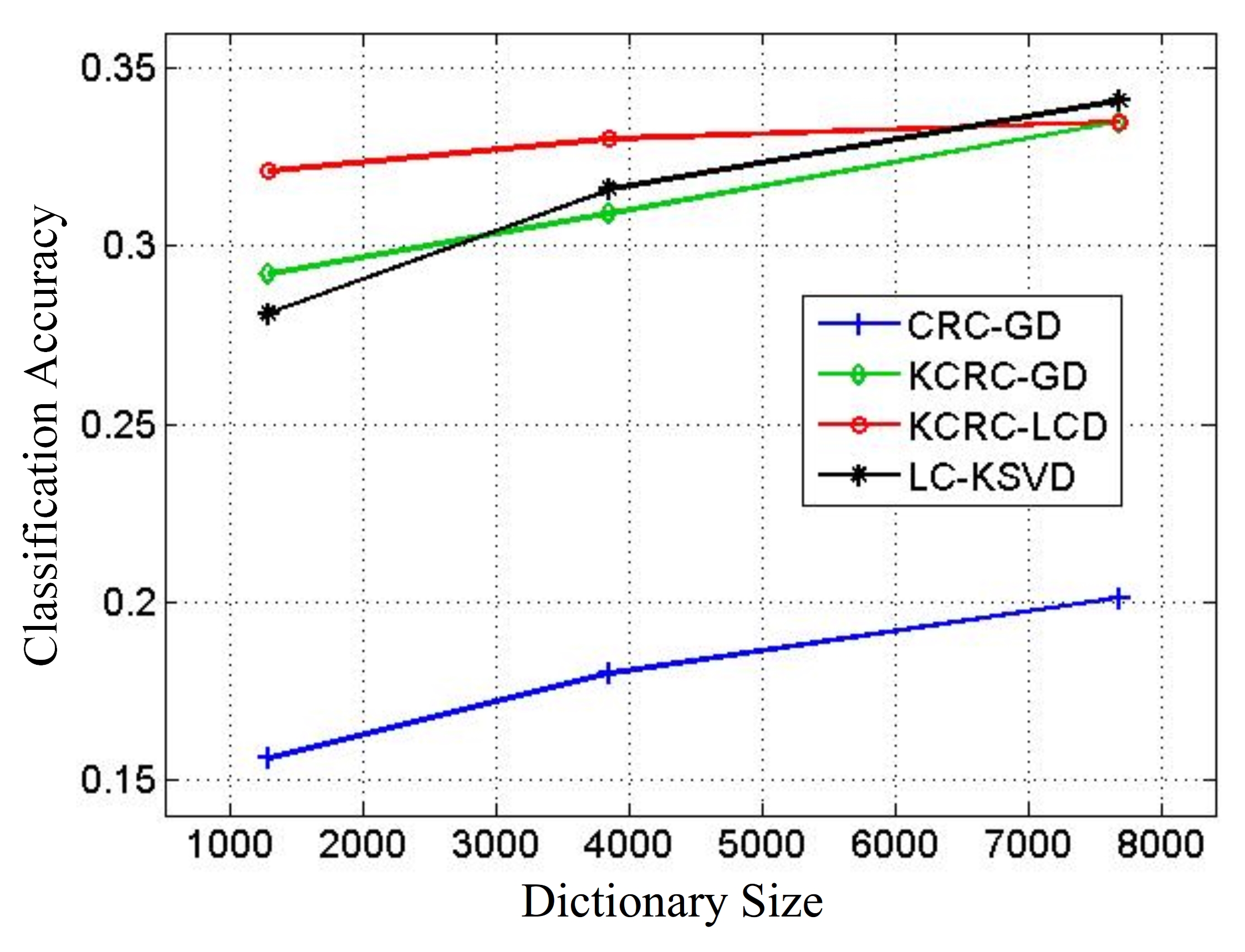}\\
  \caption{.\ \footnotesize Performance comparison with CRC-GD, KCRC-GD KCRC-LCD, and LC-KSVD \cite{jiang2013label,jiang2011learning} on Caltech256 under different dictionary size. Note that, $l_2$ distance and Eigenface features are used for similarity measure and classification. }\label{fig:cal256}
\end{figure}
The Caltech256 database \cite{griffin2007caltech} contains 30607 images of 256 categories, each category with more than 80 images. It is a very difficult visual categorization database due to the large variations in object background, pose and size. We experiment KCRC-LCD on 5, 15 and 30 training samples per category (dictionary size is 1280, 3840 and 7680 respectively). For the settings of KCRC-LCD, use the global dictionary of size 7680 (30 training samples per category) to generate the LCD and set $K$ for LCD as 1280, 3840 and 7680 for comparison. $l_2$ distance and 504-dimension Eigenface features are used for similarity measure and classification. Results in Fig. \ref{fig:cal256} shows when dictionary size is 7680, KCRC-GD, KCRC-LCD and locality constrained K-SVD (LC-KSVD) have similar classification accuracy. While LC-KSVD is slightly better, it can be observed that KCRC-LCD performs better with small dictionary size.
\subsubsection{15 Scene Categories}
\begin{figure*}[htb]
  \renewcommand{\captionlabelfont}{\footnotesize}
  \setlength{\abovecaptionskip}{0pt}
  \setlength{\belowcaptionskip}{0pt}
  \centering
  \includegraphics[width=6in]{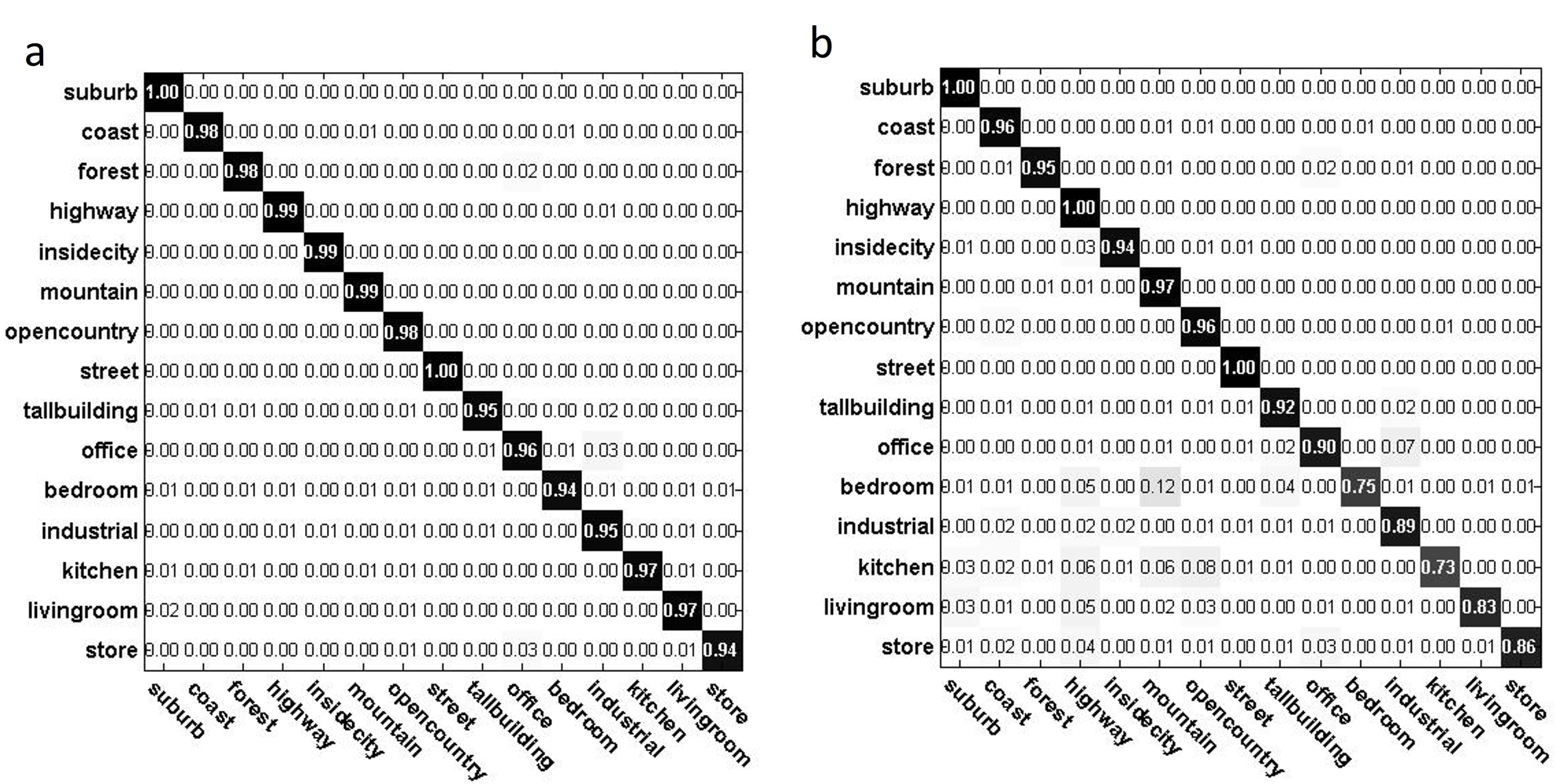}\\
  \caption{.\ \footnotesize Confusion matrices of (a) KCRC-LCD and (b) CRC-GD on the 15 scene categories database with dictionary size 450. Note that, $l_2$ distance and the spatial pyramid features are used for similarity measure and classification respectively.}\label{fig:fig5}
\end{figure*}

This database contains 15 natural scene categories such as office, kitchen and bedroom, introduced in \cite{lazebnik2006beyond}. Following the same experimental settings as \cite{jiang2013label}, we randomly select 30 images per category for training and the rest for testing. Note that, we generate the LCD with $K=450$ from a global dictionary of size 1500, similar to the training settings of LC-KSVD. Results in Table. \ref{scenetab} and Fig. \ref{fig:fig5} validate the superiority of KCRC-LCD in scenes.
\begin{table*}[htb]
  \renewcommand{\captionlabelfont}{\footnotesize}
  \setlength{\abovecaptionskip}{2pt}
  \setlength{\belowcaptionskip}{0pt}
  \centering
  \caption{.\ \footnotesize Classification results using spatial pyramid features on 15 scene categories database. Both the dictionary size and $K$ for LCD are set as 450.}\label{scenetab}
\begin{tabular}{c c|c c}\hline
Method & Accuracy(\%) & Method & Accuracy(\%)\\\hline
KCRC-LCD & 98.05 & LC-KSVD \cite{jiang2013label,jiang2011learning} & 92.94\\
KCRC-GD & 97.21 & D-KSVD \cite{zhang2010discriminative}& 89.16\\
CRC-LCD & 92.13 & KSRC \cite{zhang2012kernel} -LCD& 96.97\\
CRC-GD & 90.92 & KSRC \cite{zhang2012kernel} -GD & 95.21 \\\hline
\end{tabular}
\end{table*}
\subsection{Experiments on Running Time}
We conduct experiments on running time to evaluate the computational cost of KCRC-LCD. We use public databases including MNIST, extended Yale B, Caltech101 and 15 scene categories to perform our experiments. The detailed experimental settings are given in Table. \ref{database}. Note that, $l_2$ distance is used for similarity measure. For SRC, we use the basis pursuit (BP) algorithm to solve the $l_1$ minimization problem. Experimental results are shown in Table. \ref{time}. Compared to SRC approach, KCRC performs much faster due to its $l_2$ regularization. Constrained with locality, KCRC-LCD performs faster than KCRC-GD and CRC-GD under most circumstances.
\begin{table*}[htb]
  \renewcommand{\captionlabelfont}{\footnotesize}
  \setlength{\abovecaptionskip}{2pt}
  \setlength{\belowcaptionskip}{0pt}
  \centering
  \caption{.\ \footnotesize Experimental settings for running time test.}\label{database}
\begin{tabular}{c|c c c c c}\hline
Database & Feature Dimension & Category & Training Size & Testing Size & $K$ for LCD \\\hline
MNIST  & 784 & 10 & 500 & 10000 & 50 \\
Extended Yale B & 504 & 38 & 1216 & 1198 & 200\\
Caltech101 & 3000 & 102 & 3060 & 6084 & 510 \\
15 Scene Categories & 3000 & 15 & 1500 & 2985 & 200 \\\hline
\end{tabular}
\end{table*}
\begin{table*}[htb]
  \renewcommand{\captionlabelfont}{\footnotesize}
  \setlength{\abovecaptionskip}{2pt}
  \setlength{\belowcaptionskip}{0pt}
  \centering
  \caption{.\ \footnotesize Comparison results of average running time (ms) per classification.}\label{time}
\begin{tabular}{c |c c c c}\hline
Database & KCRC-LCD & KCRC-GD & CRC-GD & SRC-GD  \\\hline
MNIST  & 2.2618  & 3.4723 & 2.5451 & 338.69 \\
Extended Yale B & 90.434 & 367.31 & 200.48 & 587.31 \\
Caltech101 & 3016.1 & 9863.3 & 2350.9 & 19147 \\
15 Scene Categories & 246.64 & 2516.7 & 430.48 & 5943.8 \\\hline
\end{tabular}
\end{table*}

\subsection{Evaluation of The Unified Distance Measurement}
Distance metrics are of great importance in the KCRC-LCD, since they grant KCRC-LCD the scalability and discrimination power. Selecting the proper distance metric for the objects can greatly enhance the classification accuracy. Therefore, we validate the superiority of discriminative distance metrics by comparing different distance metrics in USPS, extended Yale B, MNIST and Caltech101 databases. For KCRC-LCD, we use the identity matrix as $\bm{\Psi}$. We use Euclidean distance as baseline for comparison. The USPS database contains 9288 handwirtten digits collected from mail envelops \cite{lecun1989backpropagation}. There are 7291 images for training and 2007 images for testing. This database is fairly difficult since its human error rate is 2.5\% \cite{zhang2006svm}. We apply tangent distance to construct the LCD. Specifically, tangents are attained by smoothing each image with a Gaussian kernel of width $\sigma=0.7$. Results are shown in Table. \ref{tab:DM}. For extended Yale B database, we use the same experimental settings as the previous subsection. We apply the distance metric that is proposed in \cite{ahonen2004face}. In detail, local binary pattern (LBP) histograms are extracted from divided face area and concatenated into a single feature histogram. Then $\chi^2$ distance is used to measure the similarity of different face histograms. The neighborhood for LBP operator is set as $(8,2)$ and the window size is $11\times 13$. We term the distance as LBP-$\chi^2$ distance. Results are shown in Table. \ref{tab:DM}. The MNIST database \cite{lecun1998gradient} of handwritten digits contains 60,000 samples (10 digits) for training and 10,000 for testing. We randomly select 20 samples per digit and construct a global dictionary of 200 size and test on the given 10000 samples. $K$ for LCD is set as 50 and raw pixel features are used. Results are given in Table. \ref{tab:DM}. For Caltech101 database \cite{fei2007learning}, we randomly select 30 samples per class and test on the rest (global dictionary size is 3060). $K$ for LCD is set as 500 and spatial pyramid features are used. Results are given in Table. \ref{tab:DM}.
\par
Experimental results in Table. \ref{tab:DM} show that properly selecting a good distance metric can enhance the discrimination power, and that the performance of different distance metrics can vary significantly as the distance changes (eg., Euclidean distance performs worse than Correlation distance on both USPS and Extended Yale B, but better on MNIST and Caltech101). Such variation is ubiquitous since in general every distance metric only works well under certain situations, and that the characteristics among different datasets are different due to dataset bias. Fortunately, any distance metric can be adopted into our proposed KCRC-LCD framework, showing its flexibility and generalization ability. But while our framework allows such flexibility, the unified distance measurement framework allows to by pass the troublesome process of traversing every single metric to examine its performance. In Table. \ref{tab:DM}, we can see although the results of unified distance on USPS and extended Yale B databases are not the best, but they are still very close to the optimal one. Basically, one does not need to consider the distance metrics one by one and the unified framework has automatically select the good ones, remedidating the distance metric biases. In addition, the main reason why the unified framework is not the best is because the distance metrics not complementary in this dataset (There are very few cases where other distances can complement the Tangent distance, and including other distances are essentially just poisoning the good results). Also the classification performance is pretty saturated (close to $100\%$) and the dataset itself not diverse enough. If one uses another distance that is complementary to Tangent distance and LBP-$\chi^2$ distance, we believe the classification accuracy on USPS and extended Yale B databases will be further improved. On MNIST and Caltech101 however, one could see that the combination even brings performance gain over the best single distance.

\begin{table*}[htb]
  \renewcommand{\captionlabelfont}{\footnotesize}
  \setlength{\abovecaptionskip}{2pt}
  \setlength{\belowcaptionskip}{2pt}
  \centering
  \caption{.\ \footnotesize Recognition results (\%) of different distance metrics on USPS database, MNIST database, extended Yale B database and Caltech101 database.}\label{tab:DM}
\begin{tabular}{c|c c c c}\hline
Distance Metric & USPS & Extended Yale B & MNIST & Caltech101 \\\hline
Euclidean Distance & 95.49 & 96.93 & 85.31 & 72.83\\
Manhattan Distance & 94.88 & 96.26 & 84.93 & 70.92\\
Correlation Distance & 95.57 & 97.01 & 85.17 & 72.55\\
Tangent Distance & 97.67 & N/A & N/A & N/A\\
LBP-$\chi^2$ Distance & N/A & 98.53 & N/A & N/A\\
Unified Distance & 97.17 & 98.22 & 86.52 & 73.05\\\hline
\end{tabular}
\end{table*}

\section{Conclusions}
We elaborated the KCRC approach in which kernel technique is smoothly combined with CRC. KCRC enhances the discrimination ability of CRC, making the decision boundary more reasonable. Additionally, we present a locality constrained dictionary, of which the locality is exploited to further enhance the classification performance. KCRC and LCD are mathematically linked via distance kernelization. On one hand, LCD not only helps the classifier adaptive and scalable to large databases via pruning the dictionary, but also reduce the dimensionality in kernel space, enhancing both discrimination ability and efficiency. On the other hand, kernel function makes our approach discriminative and robust to more data distribution, i.e., the same direction distribution. Furthermore, the coarse-to-fine classification strategy of KCRC-LCD is similar to the human perception process, which makes the intuition of KCRC-LCD even more appealing.
\par
We conduct comprehensive experiments to show the superiority of KCRC-LCD. Our approach yields very good classification results on various well-known public databases. While achieving high level discrimination ability, efficiency is one of the biggest merits of KCRC-LCD, which is validated in running time test. Moreover, we simulate the representation and construction of KCRC with different dimensionality reduction for kernel space, and further experiment these methods on public database. The simulation results show the discrimination ability of KCRC. Different distance metrics used in LCD are also compared to support the idea that discriminative distance metric can greatly improve the classification accuracy. We also create a toy data sets to show CRC suffers from data with the same direction distribution while KCRC perfectly overcomes such shortcoming. To sum up, tested by various experiments, KCRC is proven discriminative and efficient when combined with LCD.
\par
Possible future work includes improving the unified similarity measure model and learning the most effective kernel for KCRC-LCD instead of selecting the fixed kernel. It can be predicted that KCRC-LCD will becomes more powerful when combined with kernel learning, or even multiple kernel learning.



\bibliographystyle{elsarticle-num}
\bibliography{refs}
\end{document}